\documentclass[10pt,twocolumn,letterpaper]{article}

\usepackage[final]{cvpr} 

\usepackage{multirow}
\usepackage{colortbl}
\usepackage{bm}
\usepackage{amsthm}
\usepackage{mathtools}
\usepackage{xspace}
\usepackage{microtype}
\usepackage{placeins}

\definecolor{oursrow}{RGB}{232,240,252}

\newcommand{\method}{CHORD\xspace}
\newtheorem{proposition}{Proposition}
\newtheorem{lemma}{Lemma}

\newcommand{\E}{\mathbb{E}}
\newcommand{\KL}{\mathrm{KL}}
\newcommand{\Lpair}{\mathcal{L}_{\text{pair}}}
\newcommand{\Lcal}{\mathcal{L}_{\text{cal}}}

\newcommand{\BPI}{\mathrm{BPI}}
\newcommand{\Ibp}{\mathcal{I}_{\text{bp}}}
\newcommand{\vu}{\mathbf{u}}
\newcommand{\vh}{\mathbf{h}}
\newcommand{\va}{\mathbf{a}}
\newcommand{\vq}{\mathbf{q}}
\newcommand{\vw}{\bm{\omega}}

\definecolor{cvprblue}{rgb}{0.21,0.49,0.74}
\usepackage[pagebackref,breaklinks,colorlinks,allcolors=cvprblue]{hyperref}
\AtBeginDocument{\crefname{appendix}{Appendix}{Appendices}\crefname{subappendix}{Appendix}{Appendices}}

\def\confName{CVPR}
\def\confYear{2027}

\title{Learning What Pairs Miss for Higher-Order Multimodal Alignment}

\author{
Kunyu Zhang$^{1}$, Shenghui Feng$^{2}$, Xiangjie Shi$^{3}$, 
Zeqing Wang$^{4}$, Yingxu Wang$^{5}$\\
$^{1}$EBioMed \space
$^{2}$JLU \space
$^{3}$USTB \space
$^{4}$PolyU HK \space
$^{5}$MBZUAI
}

\begin{document}
\twocolumn[{%
\renewcommand\twocolumn[1][]{#1}%
\maketitle
\vspace{-2.8em}
\begin{center}
  \includegraphics[width=\textwidth]{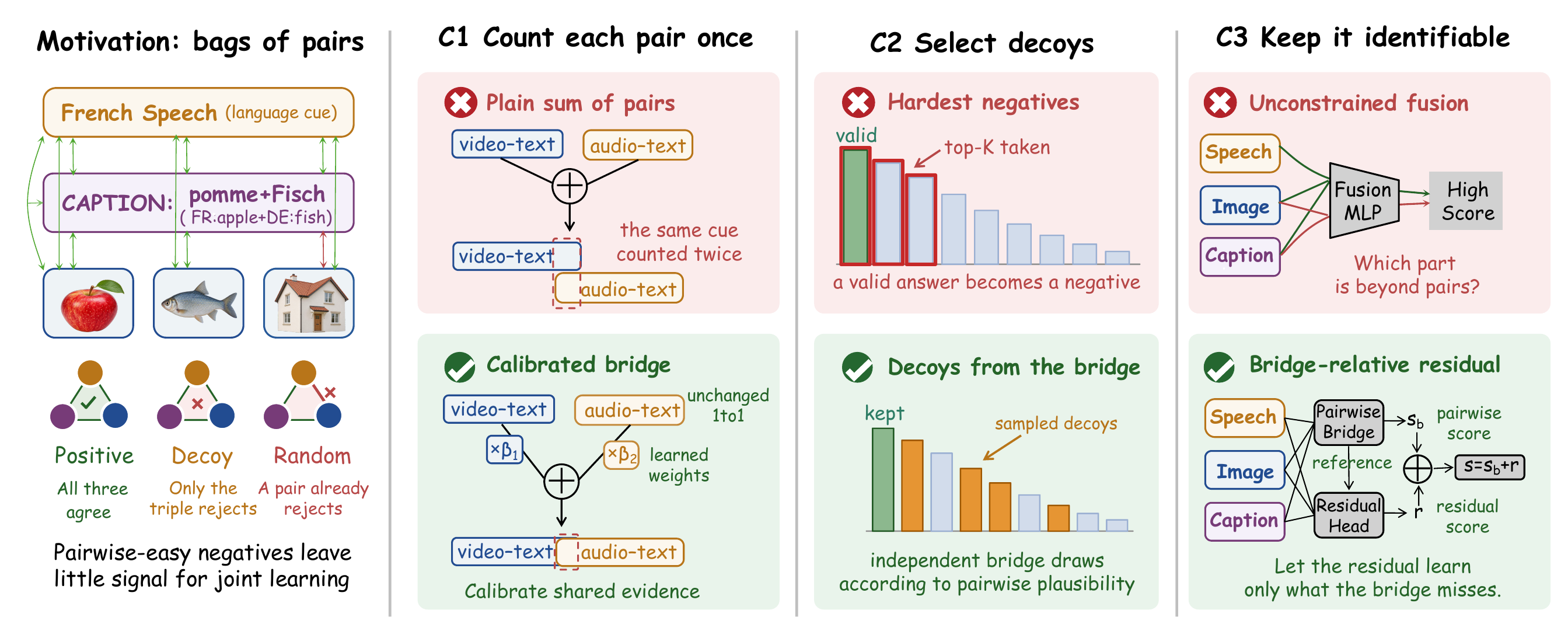}
  \vspace{-0.9em}
  \captionsetup{hypcap=false}
  \captionof{figure}{Bags of pairs and the challenges behind \method. Left: triangle edges are pairs and the face is the triple. A random negative is rejected by a single pair, whereas a decoy such as the fish passes every pair and only the triple rejects it. Right: each challenge, its naive answer (top), and our fix (bottom).}
  \label{fig:teaser}
\end{center}
\vspace{0.4em}
}]
\begin{abstract}
Contrastive learning beyond two modalities now scores the whole tuple with multilinear products, volumes, and fusion heads, critics that can represent the joint distribution. In practice these critics still behave like bags of pairs. Their negatives break at least two of the three pairs, a single pairwise check rejects them, and when most samples can be solved by pairs the higher-order term receives almost no gradient. To solve this problem, we propose \textbf{\method} (\textbf{C}ontrastive \textbf{H}igher-\textbf{O}rder \textbf{R}esiduals with \textbf{D}ecoys), which trains a residual against decoys drawn from a calibrated pairwise model, negatives that the pairwise model finds plausible, so that rejecting them often needs all three modalities. Drawing decoys costs one inner-product query over a memory bank, and the optimal residual is the log-ratio between the data and the pairwise model, so the decoy loss also bounds what that model misses. On M3-Mix, where one in ten training samples needs all three modalities, our \method improves accuracy on those samples by 21.2 percentage points over the strongest of eight baselines. \method also gives the best two-to-one results on every real dataset, and its residual improves other retrievers without retraining.\par\vspace{-12pt}
\end{abstract}

\section{Introduction}
\label{sec:intro}

Contrastive alignment is a standard way to connect modalities~\cite{wang2025protomol}: CLIP-style objectives learn one encoder per modality and score a pair with a dot product, which keeps retrieval cheap~\cite{radford2021clip,jia2021align}. With more modalities, such as the frames, sound, and transcript of a certain video, the common recipe sums pairwise losses or aligns every modality to an anchor~\cite{guzhov2022audioclip,girdhar2023imagebind,zhu2024languagebind,chen2023vast}. A recent line of work scores the whole tuple to capture information that appears only when modalities are seen together~\cite{zhang2026modeling}. For example, Symile uses a multilinear product~\cite{saporta2024symile}, GRAM a Gram volume~\cite{cicchetti2025gram}, TRIANGLE a triangle area~\cite{cicchetti2025triangle}, ConFu a fusion head~\cite{koutoupis2026confu}, and Gated Symile a gated product~\cite{rheude2026gated}. Furthermore, theory agrees that a sum of pairwise similarities cannot represent a general joint distribution over three modalities, while a weighted elementwise product can solve the problem~\cite{stuart2026expressivity}.

If expressivity is no longer the bottleneck, why are the gains of these objectives on real data small and uneven~\cite{koutoupis2026confu,ghosh2026retriangle,rheude2026gpe}? The answer lies in what the positive is contrasted against. Every negative is built by swapping in modalities from other samples, so it breaks at least two of the three pairs. When pairs carry most of the evidence, a single pairwise check rejects almost any such negative, much like a student who passes a multiple-choice test because each wrong option contradicts one line of the question. Image-text models trained against random captions show the same effect one level down and behave like bags of words~\cite{yuksekgonul2023bagofwords}. Tri-modal models behave like \emph{bags of pairs}. As shown in \cref{fig:teaser}, French speech and a caption that reads \textit{pomme} and \textit{Fisch} point to a picture of an apple. A picture of a fish is a decoy: each pair it forms looks right, and only the triple rules it out. As fewer training samples need the triple, every higher-order objective we tested falls back toward pairwise training on exactly those samples, and at one in ten it has lost about two thirds of its advantage. A contrastive critic estimates how much more likely a tuple is under the data than under independence, and this quantity splits into the part a pairwise model carries and the remainder it misses. The remainder learns only from negatives that the pairwise part cannot reject, with a gradient weighted by their share of the softmax. Pair-breaking negatives shrink this share as the pairwise evidence sharpens, so the higher-order term is starved, not unrepresentable. Training it against decoys, negatives that the pairwise model itself finds plausible, restores the signal.

Building on decoys raises the three challenges on the right of \cref{fig:teaser}. (i) Decoys need a pairwise reference that proposes plausible candidates. A plain sum of separately trained pairwise scores counts the same evidence twice when modalities repeat each other, and any change to the pairwise model risks hurting one-to-one retrieval. (ii) Decoys must come from a distribution with a clear meaning, at low cost. The hardest candidates are cheap to find but ill-defined, since a hard candidate may be a valid completion, and pushing it away teaches the model something false~\cite{robinson2021hard,chuang2020debiased}. (iii) What is learned against decoys should be identifiable. It should vanish where the pairwise model already explains the data, and its loss should measure how much that model misses, apart from the model's own errors.

To this end, we propose \textbf{\method} (\textbf{C}ontrastive \textbf{H}igher-\textbf{O}rder \textbf{R}esiduals with \textbf{D}ecoys), whose three components answer these challenges in turn. A \emph{calibrated pairwise bridge} sums pairwise similarities with constant weights fitted on the two-to-one task. The weights remove double counting, one-to-one retrieval uses the bridge unchanged, and any frozen binding model can serve as the bridge. A \emph{decoy sampler} exploits the fact that the bridge is a sum of similarities. Its distribution over a memory bank is the softmax of a single inner-product query, so drawing decoys costs one matrix-vector product. Contrasting against these decoys turns the loss into noise-contrastive estimation with the bridge as the noise distribution~\cite{gutmann2010nce,ma2018nce}, which estimates a density ratio and keeps a valid decoy at its true ratio to the bridge. A \emph{residual interaction head}, a zero-initialized weighted elementwise product of modality features, is trained only against decoys. Its optimum is the log-ratio between the data and the bridge, it stays at zero when the bridge matches the data, and its loss bounds what the bridge misses. 

Overall, our main contributions are summarized as follows:
\begin{itemize}[leftmargin=*,itemsep=1pt,topsep=2pt,parsep=0pt]
\item We identify the bags-of-pairs behavior of higher-order contrastive objectives and trace it to pair-breaking negatives, which starve the higher-order term.
\item We propose \method, which trains a residual against decoys from a calibrated pairwise bridge, with a provable optimum, unchanged one-to-one retrieval, and a loss that separates third-order information from bridge misfit.
\item Across 15 benchmarks, \method achieves the best two-to-one and fused zero-shot results on every real dataset and improves accuracy on the M3-Mix samples that need all three modalities by 21.2 percentage points over the strongest baseline. Its residual also improves other retrievers without retraining.
\end{itemize}

\begin{figure*}[t]
  \centering
  \includegraphics[width=\textwidth]{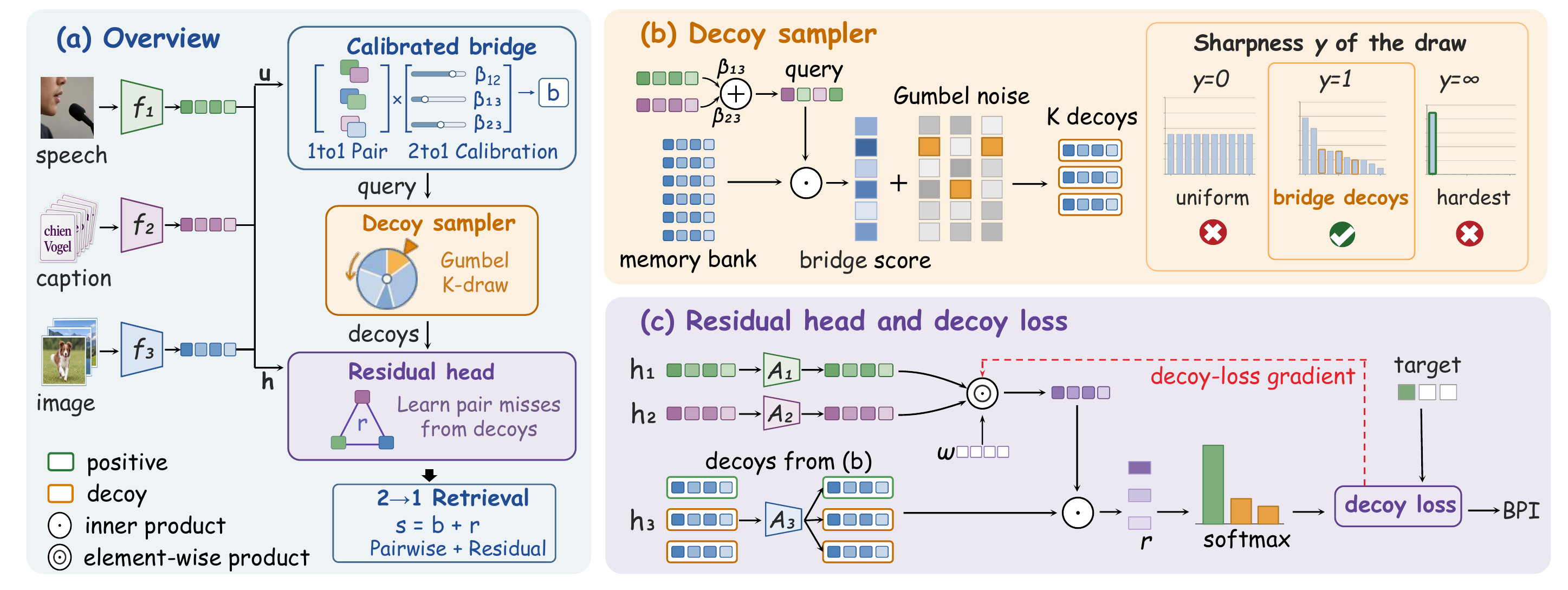}
  \vspace{-14pt}
  \caption{Overview of \method. (a) Encoders feed a calibrated bridge, which also serves one-to-one retrieval, a decoy sampler, and a residual head, and two-to-one retrieval adds the bridge and residual scores. (b) The bridge scores a memory bank with one query, and $K$ independent Gumbel-max draws pick the decoys. Sharpness $\gamma{=}1$ draws from the bridge itself, between uniform ($\gamma{=}0$) and hardest ($\gamma{\to}\infty$) draws. (c) A zero-initialized product of projected features scores the positive against the decoys, and the decoy loss does not reach the bridge.}
  \label{fig:pipeline}
  \vspace{-6pt}
\end{figure*}

\section{Related Work}
\label{sec:related}

\noindent\textbf{Contrastive Alignment of Three or More Modalities.} Pairwise and anchor-based objectives extend CLIP~\cite{radford2021clip,jia2021align} by summing losses over modality pairs or by binding all modalities to images or language~\cite{guzhov2022audioclip,girdhar2023imagebind,zhu2024languagebind,chen2023vast}. Higher-order objectives score the whole tuple. Symile maximizes a lower bound on total correlation~\cite{watanabe1960tc} with a multilinear inner product~\cite{saporta2024symile}, GRAM and TRIANGLE contrast the volume or area spanned by the embeddings~\cite{cicchetti2025gram,cicchetti2025triangle}, PMRL aligns the leading singular vector of the embedding matrix~\cite{liu2025pmrl}, and ConFu adds a fused-pair term aligned to the third modality~\cite{koutoupis2026confu}. Gated Symile makes the product robust to unreliable modalities~\cite{rheude2026gated}, OVA-IB combines one-vs-all sufficiency with an information bottleneck~\cite{li2026ovaib}, and GRAMformer scores volumes inside attention~\cite{cicchetti2026gramformer}. Stuart and Wolf prove that sums of pairwise scores cannot approximate a general joint distribution while a Hadamard-product critic can, and leave open whether training actually finds such solutions~\cite{stuart2026expressivity}. Reproductions report fragile gains~\cite{ghosh2026retriangle}, and encoder conditioning is a further bottleneck~\cite{rheude2026gpe}. A related line learns or measures interactions for supervised fusion~\cite{zhang2025mvho}, through partial information decomposition~\cite{liang2023pid,yang2026dmil}, factorized contrastive objectives~\cite{liang2023factorcl}, and synergy-aware augmentation or masking~\cite{dufumier2025comm,wen2025infmasking}. All of these methods change the critic or the representation. \method keeps a standard product critic and changes what it is contrasted against, which addresses the open question of Stuart and Wolf.

\noindent\textbf{Negatives, Noise Distributions, and Interaction Orders.} Hard-negative mining~\cite{robinson2021hard,kalantidis2020mochi} and debiasing~\cite{chuang2020debiased} reshape the negatives of bimodal contrastive learning and trade hardness against false negatives. NegCLIP adds compositional hard negatives to expose and reduce bag-of-words behavior in image-text models~\cite{yuksekgonul2023bagofwords}. Noise-contrastive estimation learns a density ratio against an arbitrary noise distribution~\cite{gutmann2010nce,ma2018nce}, and a noise distribution close to the data makes it more efficient, as in flow contrastive estimation~\cite{gao2020fce} and telescoping density-ratio estimation~\cite{rhodes2020tre}. The information-geometric hierarchy separates what pairwise couplings carry from connected higher-order information~\cite{amari2001hierarchy,schneidman2003network}. \method joins these threads: the calibrated pairwise model serves as the noise distribution, its decoys set the hardness without a tuned threshold, and the residual is tied to connected information.

\section{Methodology}
\label{sec:method}

\subsection{Preliminaries}
\label{sec:prelim}

We observe tuples $(x_1,x_2,x_3)\sim p$ of three modalities. Modality $m$ has an encoder $f_m$ with features $\vh_m=f_m(x_m)$ and a projection head $g_m$ with unit embedding $\vu_m=g_m(\vh_m)/\|g_m(\vh_m)\|$. For a target modality $k$ we write $S=\{i,j\}$ for the other two. Two-to-one retrieval ranks candidates $x_k$ by the conditional log-ratio
\begin{equation}
  \log\frac{p(x_k\mid x_S)}{p(x_k)},
  \label{eq:ratio}
\end{equation}
which InfoNCE estimates up to a term constant in $x_k$~\cite{oord2018cpc,poole2019bounds}. One-to-one retrieval uses $x_i$ in place of $x_S$. A pairwise critic such as Tri-CLIP scores a tuple by $\sum_{i\in S}\langle\vu_i,\vu_k\rangle/\tau$, and a product critic such as Symile or Hadamard-CLIP (H-CLIP) by $\vw^\top(\va_1\odot\va_2\odot\va_3)$~\cite{saporta2024symile,stuart2026expressivity}. Either way, InfoNCE contrasts the positive $x_k^+$ with targets of other tuples in the batch, drawn independently of $x_S$.

\subsection{Higher-Order Critics Behave like Bags of Pairs}
\label{sec:starve}

Write a critic as $s=b+r$, with $b$ a sum of pairwise terms and $r$ the remainder. For one query with positive $x_k^+$ and $N{-}1$ negatives, let $\Delta_n=b(x_S,x_k^+)-b(x_S,x_k^{(n)})$ be the pairwise margin of negative $n$.

\begin{proposition}[Starvation]
\label{prop:starve}
Let $\theta$ be parameters of $r$ on which $b$ does not depend, with $|r|\le B$ and $\|\nabla_\theta r\|\le G$, and let $\ell$ be the InfoNCE loss with logits $b+r$. Then
\begin{equation}
  \|\nabla_\theta\ell\|\;\le\;2G\sum_{n=1}^{N-1}\min\big\{1,\;e^{2B-\Delta_n}\big\}.
  \label{eq:starve}
\end{equation}
If negatives are drawn independently of $x_S$, then for every $\delta>0$, $\E\|\nabla_\theta\ell\|\le2G(N{-}1)\big(\pi_\delta+e^{2B-\delta}\big)$, where $\pi_\delta=P(\Delta\le\delta)$ is the probability, over queries and negatives, that a negative is pairwise plausible.
\end{proposition}

The bound holds for one query: the residual's own parameters receive gradient only from negatives that the pairwise part cannot reject. Suppose a fraction $\rho$ of queries needs the triple, such a query has $c$ pairwise-plausible targets out of $C$, as the fish in \cref{fig:teaser}, and the other queries have none. The probability at a synergy query is then about $c/C$, its average over queries is $\pi_\delta\approx\rho c/C$, and the bound on the average gradient scales as $N\rho c/C$ for any form of $r$.

\subsection{\method}
\label{sec:decoy}

In this subsection, we introduce the design of each component of our \method. The structure of three components of \method are shown in \Cref{fig:pipeline}.

\noindent\textbf{Calibrated Pairwise Bridge.} For target $k$ the bridge is
\begin{equation}
  b_k(x_S,x_k)=\big\langle\vq_k(x_S),\vu_k\big\rangle,
  \label{eq:bridge}
\end{equation}
where $\vq_k(x_S)=\tau_b^{-1}\sum_{i\in S}\beta_{ik}\vu_i$, with one weight per pair of modalities, $\beta_{ik}=\beta_{ki}\ge0$, and a temperature $\tau_b$. The embeddings are trained with the pairwise InfoNCE loss $\Lpair$, and $\beta$ and $\tau_b$ with a two-to-one InfoNCE loss $\Lcal$ on the logits $b_k$. Calibration matters when modalities repeat each other: for two copies, both pairwise scores estimate the same dependence and their plain sum counts it twice. Because the weights are symmetric, the three conditionals of the bridge are those of one joint pairwise model $q_b(x)\propto\prod_m p(x_m)\exp\sum_{i<j}\beta_{ij}\langle\vu_i,\vu_j\rangle/\tau_b$, so one residual can correct all three directions. 
For one-to-one retrieval, we rank candidates directly by the inner product $\langle\vu_i,\vu_k\rangle$.

\noindent\textbf{Decoys from the Bridge.} The bridge induces $q_b(x_k\mid x_S)\propto p(x_k)\exp b_k(x_S,x_k)$. A memory bank $\{\vu_k^{(n)}\}_{n=1}^{Q}$ of past targets~\cite{he2020moco} holds samples of $p(x_k)$, so on the bank $q_b$ is a softmax of $Q$ inner products with one query vector,
\begin{equation}
  q_b(n\mid x_S)=\frac{\exp\big(\gamma\langle\vq_k(x_S),\vu_k^{(n)}\rangle\big)}{\sum_{n'=1}^{Q}\exp\big(\gamma\langle\vq_k(x_S),\vu_k^{(n')}\rangle\big)},
  \label{eq:sampler}
\end{equation}
with $\gamma=1$ in the default setting. We draw $K$ decoys independently from this distribution, each with its own Gumbel perturbation~\cite{kool2019gumbel}, at the cost of one $Q\times d$ matrix-vector product per query. Nothing is removed from the bank, whose distribution converges to $q_b$ as $Q$ grows, whereas drawing without replacement or removing the positive biases the estimator when $q_b$ is concentrated. The sharpness $\gamma$ moves from uniform negatives ($\gamma=0$) toward draws concentrated on the single highest-scoring candidate ($\gamma\to\infty$), as \cref{fig:pipeline}b sketches. As a separate baseline, deterministic top-$K$ mining selects the $K$ highest-scoring distinct candidates, whereas all draws in the limit select the same candidate.


\noindent\textbf{Residual and Decoy Loss.} The residual is a weighted elementwise product of projected features,
\begin{equation}
  r(x_1,x_2,x_3)=\vw^\top\big(\va_1\odot\va_2\odot\va_3\big),
  \label{eq:residual}
\end{equation}
with $\va_m=\mathrm{LN}(A_m\vh_m)\in\mathbb{R}^{R}$, the critic class of H-CLIP~\cite{stuart2026expressivity}. We initialize $\vw=\mathbf{0}$, and one head serves all targets. For target $k$ the decoy loss is
\begin{equation}
  \mathcal{L}_{\text{dec},k}=-\E\log\frac{e^{r(x_S,x_k^+)}}{e^{r(x_S,x_k^+)}+\sum_{d=1}^{K}e^{r(x_S,x_k^{(d)})}},
  \label{eq:decoyloss}
\end{equation}
with the decoys drawn above. This is ranking noise-contrastive estimation~\cite{ma2018nce} with proposal $q_b$, whose noise-corrected logits $b_k+r-\log(q_b/p(x_k))$ reduce to $r$ up to a query constant. For other sharpness values the corrected logits are $r+(1-\gamma)\,b_k$ with $b_k$ held fixed, so every $\gamma$ targets the same residual and changes only how hard the decoys are. The case $\gamma=0$, a uniform proposal from the bank with logits $b_k+r$, is the matched uniform control.

\noindent\textbf{Training and Inference.} The objective is $\mathcal{L}=\Lpair+\Lcal+\lambda\sum_k\mathcal{L}_{\text{dec},k}$ with $\lambda=1$, and all terms are optimized together. At $\gamma=1$ the noise-corrected logits of the decoy loss are $r$ alone, so the decoy loss never reaches the projection heads or $\beta$. Decoys are drawn without gradient, so within each step the proposal is a fixed $q_b$. We train with frozen encoders, with trainable heads, or end to end, where the decoy loss also updates the shared encoders. The two-to-one retrieval score of a candidate is formulated as
\begin{equation}
  s_k(x_S,x_k)=b_k(x_S,x_k)+r(x_S,x_k),
  \label{eq:score}
\end{equation}
one inner product over $d{+}R$ dimensions between $[\vq_k;\,\vw\odot\va_i\odot\va_j]$ and the precomputable $[\vu_k;\,\va_k]$. On held-out data we report $\BPI_K=\log(K{+}1)-\tfrac13\sum_k\mathcal{L}_{\text{dec},k}$.

\begin{table}[t]
\centering
\caption{Top-1 accuracy (\%) on controlled benchmarks. XOR-Mix and M3-Mix
keep a pairwise path for most samples and need the triple for a fraction $\rho{=}0.1$.
Pairwise training reaches 50\% on those samples (pairwise reference, \cref{supp:ref}). The best results are shown in \textbf{bold}, and the second-best results are \underline{underlined}.}
\label{tab:controlled}
\footnotesize
\setlength{\tabcolsep}{2.1pt}
\begin{tabular}{@{}l ccc cccc@{}}
\toprule
 & \multicolumn{3}{c}{Synthetic} & \multicolumn{4}{c}{Symile-M3} \\
\cmidrule(lr){2-4}\cmidrule(l){5-8}
Method & XOR & XOR-Mix & XNOR & $w{=}2$ & $w{=}5$ & $w{=}10$ & Mix \\
\midrule
Tri-CLIP & 3.8 & 50.1 & 24.4 & 47.5 & 18.8 & 9.6 & 48.6 \\
Symile & \textbf{100.0} & 64.2 & 33.2 & 93.8 & 92.0 & 88.3 & 66.2 \\
Gated Symile & \underline{99.8} & \underline{66.0} & \textbf{87.1} & 94.1 & 92.0 & 88.6 & 67.0 \\
H-CLIP & \textbf{100.0} & 65.3 & 41.6 & \underline{94.4} & \underline{92.3} & \underline{88.9} & \underline{68.4} \\
GRAM & 13.8 & 50.3 & 48.3 & 48.1 & 19.3 & 9.9 & 48.9 \\
TRIANGLE & 12.9 & 50.6 & 60.7 & 48.4 & 19.6 & 10.1 & 49.2 \\
PMRL & 9.6 & 50.2 & 45.2 & 47.8 & 19.0 & 9.7 & 48.8 \\
ConFu & \textbf{100.0} & 61.0 & 52.7 & 92.6 & 89.8 & 85.1 & 63.1 \\
\rowcolor{oursrow}\textbf{CHORD} & \textbf{100.0} & \textbf{97.9} & \underline{84.6} & \textbf{94.8} & \textbf{92.6} & \textbf{89.4} & \textbf{89.6} \\
\bottomrule
\end{tabular}
\end{table}

\begin{table*}[t]
\centering
\caption{Retrieval on MultiBench (Recall@10, \%) and the decoy probe (Decoy@16, top-1 \%, chance 6.25). 1$\to$1 averages six directions and 2$\to$1 three targets. n/a: the critic does not score pairs. The best results are shown in \textbf{bold}, and the second-best results are \underline{underlined}.}
\label{tab:multibench}
\footnotesize
\setlength{\tabcolsep}{3.3pt}
\begin{tabular}{@{}l ccc ccc ccc ccc@{}}
\toprule
 & \multicolumn{3}{c}{MOSI} & \multicolumn{3}{c}{MOSEI} & \multicolumn{3}{c}{UR-FUNNY} & \multicolumn{3}{c}{MUStARD} \\
\cmidrule(lr){2-4}\cmidrule(lr){5-7}\cmidrule(lr){8-10}\cmidrule(l){11-13}
Method & 1$\to$1 & 2$\to$1 & Decoy@16 & 1$\to$1 & 2$\to$1 & Decoy@16 & 1$\to$1 & 2$\to$1 & Decoy@16 & 1$\to$1 & 2$\to$1 & Decoy@16 \\
\midrule
Tri-CLIP & \underline{22.7} & \underline{23.1} & 8.4 & \underline{6.8} & \underline{8.9} & 8.0 & \underline{8.0} & 12.6 & 7.7 & 41.8 & 59.0 & 9.9 \\
Symile & \textcolor{gray}{n/a} & 17.0 & 9.6 & \textcolor{gray}{n/a} & 7.1 & 8.6 & \textcolor{gray}{n/a} & 11.8 & 9.3 & \textcolor{gray}{n/a} & 46.5 & 13.1 \\
Gated Symile & \textcolor{gray}{n/a} & 18.4 & \underline{10.1} & \textcolor{gray}{n/a} & 7.6 & \underline{8.9} & \textcolor{gray}{n/a} & 12.3 & \underline{9.8} & \textcolor{gray}{n/a} & 49.3 & \underline{13.9} \\
H-CLIP & \textcolor{gray}{n/a} & 18.1 & 9.9 & \textcolor{gray}{n/a} & 7.4 & 8.8 & \textcolor{gray}{n/a} & 12.1 & 9.6 & \textcolor{gray}{n/a} & 48.1 & 13.4 \\
GRAM & 6.0 & 11.7 & 7.4 & 2.9 & 5.2 & 7.0 & 2.4 & 3.6 & 6.9 & 35.1 & 60.2 & 9.8 \\
TRIANGLE & \textcolor{gray}{n/a} & 7.5 & 7.9 & \textcolor{gray}{n/a} & 5.8 & 7.3 & \textcolor{gray}{n/a} & 3.6 & 7.2 & \textcolor{gray}{n/a} & 50.3 & 10.6 \\
PMRL & 17.9 & 21.6 & 7.6 & 5.9 & 7.8 & 7.4 & 6.9 & 10.8 & 7.0 & 39.6 & 57.1 & 9.9 \\
ConFu & 19.6 & 19.8 & 9.3 & 6.3 & 8.4 & 8.5 & 7.5 & \underline{13.5} & 9.0 & \textbf{45.4} & \underline{62.7} & 12.0 \\
\rowcolor{oursrow}\textbf{CHORD} & \textbf{22.8} & \textbf{23.7} & \textbf{11.9} & \textbf{7.0} & \textbf{9.4} & \textbf{11.3} & \textbf{8.3} & \textbf{14.5} & \textbf{14.4} & \underline{43.6} & \textbf{64.4} & \textbf{18.3} \\
\bottomrule
\end{tabular}
\end{table*}

\subsection{What the Residual Learns}
\label{sec:theory}

\begin{proposition}[Optimum of the decoy loss]
\label{prop:consistency}
Let the bridge weights be symmetric and the $K$ decoys be drawn independently from $q_b(\cdot\mid x_S)$. (i) Over all measurable $r$, $\mathcal{L}_{\text{dec},k}$ is minimized by
\begin{equation}
  r^\star(x_S,x_k)=\log\frac{p(x_k\mid x_S)}{q_b(x_k\mid x_S)}+c_k(x_S),
  \label{eq:rstar}
\end{equation}
and $\log p(x)/q_b(x)$ attains this for all three targets at once. (ii) $b_k+r^\star$ equals \cref{eq:ratio} up to a term constant in $x_k$. (iii) For every $r$ and $K$, $\BPI_K\le\Ibp(q_b):=\tfrac13\sum_k\E_{x_S}\KL\big(p(\cdot\mid x_S)\,\|\,q_b(\cdot\mid x_S)\big)$, and $r^\star$ is constant in $x_k$ if and only if $\Ibp(q_b)=0$.
\end{proposition}

\begin{proposition}[No harm with frozen encoders]
\label{prop:noharm}
(i) With frozen encoders, one-to-one retrieval is that of the bridge, whatever the residual learns. (ii) If $p(\cdot\mid x_S)=q_b(\cdot\mid x_S)$, then $r^\star$ is constant and \method ranks two-to-one candidates as the bridge does. (iii) With $\vw=\mathbf{0}$ at initialization, training starts from the bridge.
\end{proposition}

The propositions concern the minimizer over all measurable residuals, with proofs in the supplementary material.

\begin{proposition}[What the bridge misses]
\label{prop:decomp}
Let $q_2^\star$ be the maximum-entropy distribution with the pairwise marginals of $p$, and assume that it exists and that the divergences below are finite. Since $q_b$ lies in the pairwise exponential family,
\begin{equation}
  \Ibp(q_b)=I^{(3)}+M(q_b),\qquad I^{(3)}=\KL\big(p\,\|\,q_2^\star\big),
  \label{eq:decomp}
\end{equation}
where $I^{(3)}$ is the connected information of order three~\cite{amari2001hierarchy,schneidman2003network} and $M(q_b)=\tfrac13\sum_k\E_{q_2^\star}\KL\big(q_2^\star(\cdot\mid x_S)\,\|\,q_b(\cdot\mid x_S)\big)\ge0$ is the pairwise misfit of the bridge.
\end{proposition}

The bound $\Ibp$ equals the connected information only when the bridge reaches $q_2^\star$, and with the optimal residual BPI approaches it from below as $K$ grows. The same holds pointwise: the residual targets $\log p/q_b$, which differs from $\log p/q_2^\star$ by $\log q_2^\star/q_b$, so the two coincide only for a bridge without misfit. On discrete synthetic data we compute $I^{(3)}$ exactly. On real data, a pairwise residual $\sum_{i<j}\vw_{ij}^\top(\va_i\odot\va_j)$ of the same rank, trained on the same decoys, absorbs most of the misfit, so $\Delta\BPI=\BPI_K-\BPI_{\text{pair}}$ serves as a diagnostic that tracks $I^{(3)}$ without bounding it. At initialization, the decoy softmax assigns a total mass of $K/(K{+}1)$ to the decoys whatever the pairwise margins, so large margins no longer shrink the weight on the negatives that train the residual.


\section{Experiments}
\label{sec:exp}

\noindent\textbf{Datasets.} We use 15 benchmarks in three groups. The controlled ones are XOR~\cite{saporta2024symile}, our XOR-Mix, Synthetic-XNOR~\cite{rheude2026gated}, and Symile-M3~\cite{saporta2024symile}. On XOR pairs carry no information, and on XOR-Mix they narrow the target to two candidates while a fraction $\rho$ of the samples needs both inputs to choose. On Symile-M3 the caption names $w$ classes and the speech selects one, and in its variant M3-Mix ($w{=}2$) only a fraction $\rho$ of the captions name two classes. The affective ones are MOSI~\cite{zadeh2016mosi}, MOSEI~\cite{zadeh2018mosei}, UR-FUNNY~\cite{hasan2019urfunny}, and MUStARD~\cite{castro2019mustard} from MultiBench~\cite{liang2021multibench}. The audio-visual ones are AV-MNIST~\cite{vielzeuf2018centralnet}, SSW60~\cite{vanhorn2022ssw60}, VB100~\cite{ge2016vb100}, MSR-VTT~\cite{xu2016msrvtt}, VATEX~\cite{wang2019vatex}, and YouCook2~\cite{zhou2018youcook2}. SSW60 and VB100 test zero-shot classification, MSR-VTT and VATEX text-to-video retrieval, YouCook2 zero-shot transfer from MSR-VTT, and Vision\&Touch~\cite{lee2019visiontouch} three to five modalities.

\noindent\textbf{Baselines.} We compare with Tri-CLIP, Symile~\cite{saporta2024symile}, Gated Symile~\cite{rheude2026gated}, H-CLIP~\cite{stuart2026expressivity}, GRAM~\cite{cicchetti2025gram}, TRIANGLE~\cite{cicchetti2025triangle}, PMRL~\cite{liu2025pmrl}, and ConFu~\cite{koutoupis2026confu}. All methods share encoders, heads, and embedding size, and each is evaluated with its own score. Tri-CLIP sums the pairwise InfoNCE losses in training and the two cosines for two-to-one retrieval.

\noindent\textbf{Implementation Details.} MultiBench uses the encoders of ConFu~\cite{koutoupis2026confu}, SSW60 and VB100 follow its Bird-MML pretraining, and the video benchmarks use frozen LanguageBind encoders~\cite{zhu2024languagebind}. We use $K{=}32$ decoys, a bank of $Q{=}16{,}384$ targets per modality, $\gamma{=}1$, and $\lambda{=}1$ everywhere, with the optimizers and schedules of the respective baselines. We report top-1 accuracy on controlled data, Recall@10 on MultiBench, Recall@1 for text-to-video retrieval, and zero-shot accuracy for fused classification. The probe \emph{Decoy@16} is a pairwise-guided stress test: 15 decoys per query come from a judge bridge trained with another seed and pass every pairwise check, a filter that training never uses, and we report top-1 accuracy among 16 candidates (chance 6.25\%). BPI uses $K{=}64$ independent decoys. Ablation and sensitivity studies use one dataset from each table of the main results, M3-Mix, MUStARD, and MSR-VTT. All numbers are means over five seeds. The supplementary material gives the remaining protocols, per-task results, and controls.

\subsection{Main Results}
\label{sec:main}

\begin{figure*}[t]
  \centering
  \includegraphics[width=\textwidth]{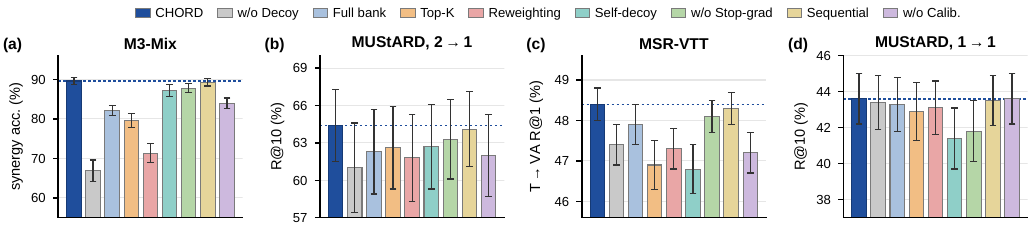}
  \vspace{-16pt}
  \caption{Ablation of \method on M3-Mix, MUStARD (two-to-one and one-to-one), and MSR-VTT. Error bars show the standard deviation over five seeds, and the dashed line marks the full model.}
  \label{fig:ablation}
  \vspace{-8pt}
\end{figure*}

\begin{table}[t]
\centering
\caption{Audio-visual tasks. Left: text$\to$(video, audio) retrieval, R@1 (\%), with frozen
LanguageBind encoders. YC2 is zero-shot transfer from MSR-VTT. Right: fused audio-visual zero-shot accuracy (\%).}
\label{tab:audiovisual}
\footnotesize
\setlength{\tabcolsep}{2.4pt}
\begin{tabular}{@{}l ccc ccc@{}}
\toprule
 & \multicolumn{3}{c}{Retrieval (T$\to$VA)} & \multicolumn{3}{c}{Zero-shot (A+V)} \\
\cmidrule(lr){2-4}\cmidrule(l){5-7}
Method & MSR-VTT & VATEX & YC2 & SSW60 & VB100 & AV-MN. \\
\midrule
\textcolor{gray}{Best single} & \textcolor{gray}{46.0} & \textcolor{gray}{69.5} & \textcolor{gray}{12.4} & \textcolor{gray}{70.2} & \textcolor{gray}{20.5} & \textcolor{gray}{64.8} \\
\midrule
Tri-CLIP & 46.8 & 70.4 & \underline{12.6} & 70.3 & \underline{19.2} & 69.4 \\
Symile & 44.9 & 67.9 & 10.9 & 60.4 & 13.1 & 71.0 \\
Gated Symile & 45.6 & 68.8 & 11.2 & 63.1 & 17.2 & \underline{71.5} \\
H-CLIP & 45.1 & 68.1 & 10.8 & 61.0 & 13.9 & 71.3 \\
GRAM & 47.3 & 71.2 & 11.4 & 57.1 & 7.9 & 64.1 \\
TRIANGLE & \underline{47.8} & \underline{71.8} & 10.7 & 64.2 & 11.8 & 65.1 \\
PMRL & 47.1 & 71.0 & 11.8 & 66.2 & 15.1 & 66.0 \\
ConFu & 47.2 & 71.3 & 11.6 & \underline{71.2} & 18.3 & 70.9 \\
\rowcolor{oursrow}\textbf{CHORD} & \textbf{48.4} & \textbf{72.4} & \textbf{12.9} & \textbf{72.3} & \textbf{20.9} & \textbf{71.9} \\
\bottomrule
\end{tabular}
\end{table}

\begin{table}[t]
\centering
\caption{One-to-one and two-to-one retrieval (R@10, \%) of \method under the three training modes.}
\label{tab:modes}
\footnotesize
\setlength{\tabcolsep}{3pt}
\begin{tabular}{@{}l cc cc@{}}
\toprule
 & \multicolumn{2}{c}{MOSI} & \multicolumn{2}{c}{MUStARD} \\
\cmidrule(lr){2-3}\cmidrule(l){4-5}
Mode & 1$\to$1 & 2$\to$1 & 1$\to$1 & 2$\to$1 \\
\midrule
Tri-CLIP (reference) & 22.7 & 23.1 & 41.8 & 59.0 \\
Frozen encoders & 22.7 & 23.4 & 41.8 & 62.6 \\
Trainable heads only & 22.7 & 23.5 & 41.8 & 63.2 \\
\rowcolor{oursrow}End to end, shared encoders & 22.8 & 23.7 & 43.6 & 64.4 \\
\bottomrule
\end{tabular}
\end{table}

As shown in \cref{tab:controlled}, expressivity and training signal come apart. On pure synergy, as in XOR and Symile-M3, product critics already do well and \method matches them, while geometric objectives stay at the pairwise level~\cite{koutoupis2026confu}. Once most samples have a pairwise path, as in XOR-Mix and M3-Mix, every baseline falls far below its accuracy on pure synergy and even below the best additive decision rule, whereas \method keeps most of its accuracy. The critics that solve XOR are the ones that fail here, so expressivity is not the issue. On Synthetic-XNOR, Gated Symile remains best and \method second, so gating complements decoys. Results on real data are shown in \cref{tab:multibench,tab:audiovisual}. \method gives the best two-to-one retrieval on all four MultiBench datasets and the highest Decoy@16 accuracy on every dataset, and its one-to-one retrieval is never below that of Tri-CLIP. As shown in \cref{tab:modes}, one-to-one retrieval equals that of the pairwise model with frozen encoders or trainable heads, and end to end it improves slightly. \method also leads on all six audio-visual columns. On VB100, where audio is weakly informative, every fused baseline falls below the visual model alone, whereas \method stays above it because calibration down-weights the weak pair. Under the shift to YouCook2, product and geometric critics drop below the summed cosines, while \method stays close to its bridge, since a residual trained on MSR-VTT adds little on a new domain.

\subsection{Ablation and Sensitivity Study}
\label{sec:ablation}

\begin{table}[t]
\centering
\caption{Residual families and proposals on one calibrated bridge, with the same bank and $K{=}32$. Uniform and decoy columns use noise-corrected logits and share the same target residual. M3-Mix reports accuracy on samples that need the triple, and MUStARD reports 2$\to$1 R@10 (\%). Hadamard with decoys is CHORD.}
\label{tab:negatives}
\footnotesize
\setlength{\tabcolsep}{2.5pt}
\begin{tabular}{@{}l ccc ccc@{}}
\toprule
 & \multicolumn{3}{c}{M3-Mix} & \multicolumn{3}{c}{MUStARD} \\
\cmidrule(lr){2-4}\cmidrule(l){5-7}
Residual & Unif. & Top-$K$ & Decoys & Unif. & Top-$K$ & Decoys \\
\midrule
Pairwise (control) & 48.8 & 49.0 & 49.3 & 61.3 & 61.5 & 61.9 \\
\rowcolor{oursrow}Hadamard (H-CLIP) & \textbf{66.9} & \textbf{79.6} & \textbf{89.6} & \underline{61.4} & \textbf{62.6} & \textbf{64.4} \\
Multilinear (Symile) & 65.0 & 77.8 & 88.3 & 61.0 & 62.0 & 63.6 \\
Gated (Gated Symile) & \underline{66.1} & \underline{78.6} & \underline{88.9} & 61.2 & 62.3 & \underline{63.9} \\
Fusion MLP (ConFu) & 62.4 & 75.1 & 86.8 & \textbf{61.5} & \underline{62.4} & 63.2 \\
Volume (GRAM) & 49.3 & 52.6 & 55.0 & 60.6 & 61.0 & 61.8 \\
Area (TRIANGLE) & 49.5 & 53.1 & 55.6 & 60.3 & 60.8 & 61.6 \\
\bottomrule
\end{tabular}
\end{table}

\begin{table}[t]
\centering
\caption{Decoy@16 (top-1 \%) with judges from three pairwise critic families on UR-FUNNY (URF), MUStARD (MUS), and SSW60 (SSW).}
\label{tab:judges}
\footnotesize
\setlength{\tabcolsep}{1.9pt}
\begin{tabular}{@{}l ccc ccc ccc@{}}
\toprule
 & \multicolumn{3}{c}{Dot product} & \multicolumn{3}{c}{Bilinear} & \multicolumn{3}{c}{Pairwise MLP} \\
\cmidrule(lr){2-4}\cmidrule(lr){5-7}\cmidrule(l){8-10}
Method & URF & MUS & SSW & URF & MUS & SSW & URF & MUS & SSW \\
\midrule
Tri-CLIP & 7.7 & 9.9 & 9.2 & 8.0 & 10.2 & 9.5 & 7.8 & 10.0 & 9.4 \\
Symile & 9.3 & 13.1 & 10.8 & 9.2 & 12.9 & 10.7 & 9.0 & 12.5 & 10.5 \\
Gated Symile & \underline{9.8} & \underline{13.9} & 11.3 & \underline{9.7} & \underline{13.6} & 11.2 & \underline{9.5} & \underline{13.3} & 11.0 \\
ConFu & 9.0 & 12.0 & \underline{12.1} & 8.9 & 11.8 & \underline{11.9} & 8.7 & 11.5 & \underline{11.7} \\
\rowcolor{oursrow}\textbf{CHORD} & \textbf{14.4} & \textbf{18.3} & \textbf{14.5} & \textbf{13.6} & \textbf{17.2} & \textbf{13.9} & \textbf{12.9} & \textbf{16.4} & \textbf{13.4} \\
\bottomrule
\end{tabular}
\end{table}

\begin{figure*}[t]
  \centering
  \includegraphics[width=\textwidth]{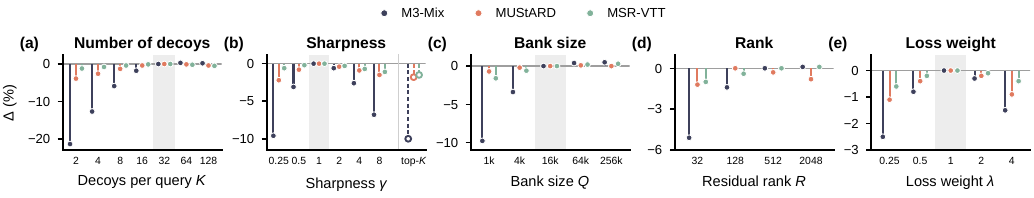}
  \vspace{-16pt}
  \caption{Sensitivity of \method: change from the default (shaded) with (a) the number of decoys, (b) their sharpness $\gamma$, with top-$K$ mining shown apart (hollow), (c) the bank size, (d) the residual rank, whose default is 128 on MUStARD and 512 otherwise, and (e) the loss weight. Standard deviations over five seeds stay below 0.5\%.}
  \label{fig:sensitivity}
  \vspace{-8pt}
\end{figure*}

The ablation results are shown in \cref{fig:ablation}. Four variants change the negatives of the residual: w/o Decoy uses uniform in-batch negatives, Full bank uses every bank target, Top-K takes the $K$ highest bridge scores, and Reweighting applies hardness reweighting~\cite{robinson2021hard} to uniform negatives. Three variants train the bridge and the residual together in other ways: w/o Stop-grad keeps the logits at $r$ but lets the decoy loss update the bridge, Self-decoy draws the decoys from $b{+}r$ and contrasts them with $b{+}r$, and Sequential trains the bridge first and the residual afterwards. Finally, w/o Calib.\ sums the pairwise scores with equal weights. Without decoys, accuracy on M3-Mix falls to the level of H-CLIP, so the same model trained against ordinary negatives starves its residual. Full bank recovers part of the gap at a far higher cost, and Top-K and Reweighting trail \method on M3-Mix and also lose on MSR-VTT, where near-duplicate captions turn the hardest candidates into false negatives. Letting the decoy loss move the bridge, in w/o Stop-grad and Self-decoy, lowers one-to-one retrieval on MUStARD because the pairwise geometry shifts, and Self-decoy also loses on MSR-VTT for the same reason as Top-K. Sequential matches \method but needs a second run. Without calibration, redundant evidence is counted twice, the decoys become too sharp, and the accuracy of w/o Calib drops on all three datasets.

The sensitivity results are shown in \cref{fig:sensitivity}, where accuracy saturates at about 32 decoys per query. With the logit correction, every finite $\gamma$ targets the same residual, so the sweep over $\gamma$ changes only how hard the decoys are: accuracy peaks at the bridge itself, $\gamma{=}1$, and falls for sharper draws, which concentrate on a few of the highest-scoring candidates. Deterministic top-$K$ mining, plotted apart, is worse than every sharper setting of $\gamma$. Banks larger than the default add little, the rank matters most on M3-Mix, where the synergy is hardest to represent, ranks above the default slightly overfit the small MUStARD, and loss weights between 0.5 and 2 barely change accuracy.

\subsection{The Bottleneck Is the Negatives}
\label{sec:negatives}

\begin{figure}[t]
  \centering
  \includegraphics[width=\linewidth]{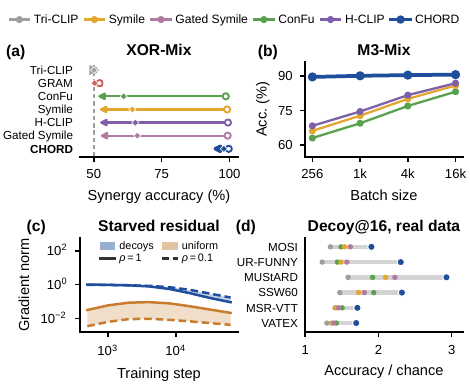}
  \vspace{-14pt}
  \caption{Diagnosing bags of pairs. (a) XOR-Mix accuracy on synergy samples as $\rho$ falls from 1 (circle) via 1/10 (diamond) to 1/50 (arrow). (b) M3-Mix accuracy versus batch size with uniform negatives (\method: 32 decoys at every size). (c) Normalized gradient norm reaching the residual on M3-Mix, where each band spans $\rho{=}1$ to $\rho{=}0.1$. (d) Decoy@16 accuracy on six real datasets as a multiple of chance.}
  \label{fig:diagnosis}
  \vspace{-8pt}
\end{figure}

If the critic were the bottleneck, changing the negatives would not matter. In \cref{tab:negatives}, seven residual families are trained on one calibrated bridge, and the uniform and decoy columns differ only in the proposal. Decoys lift the Hadamard, multilinear, gated, and fusion critics by a wide margin, so the sampler also upgrades Symile and ConFu. Volume and area critics gain much less in this setting. The pairwise residual gains almost nothing on M3-Mix and little on MUStARD, so most of the MUStARD gain comes from the interaction term, not from a repaired bridge. As shown in \cref{fig:diagnosis}a, every baseline decays toward the pairwise level as the share $\rho$ of synergy samples shrinks. As shown in \cref{fig:diagnosis}b, larger batches help only slowly, and uniform negatives stay below \method even with a batch 64 times larger, consistent with a starvation bound linear in $N$. As shown in \cref{fig:diagnosis}c, the gradient reaching the residual under uniform negatives shrinks as synergy samples become rare and stays orders of magnitude below its value under decoys. On real data, \cref{fig:diagnosis}d shows that \method is the hardest to fool on Decoy@16 on every dataset, with the largest margins on UR-FUNNY and MUStARD.
The bilinear and MLP judges in \cref{tab:judges} reproduce the full ranking obtained with the default bridge judge. Under both judges, \method retains a clear lead, though by a smaller margin.

\subsection{What the Residual Learns}
\label{sec:bpi}

\begin{table}[t]
\centering
\caption{Sum, bridge, and \method on real data, ordered by $\Delta$BPI. Res.: gain of the residual over the bridge, mean $\pm$ std over seeds. Plug-in: mean gain when the same residual is added to the eight baselines without retraining them.}
\label{tab:decomp}
\footnotesize
\setlength{\tabcolsep}{2.3pt}
\begin{tabular}{@{}l ccc cc c@{}}
\toprule
Dataset & Sum & Bridge & \method & Res. & Plug-in & $\Delta$BPI \\
\midrule
MUStARD & 59.0 & 61.2 & 64.4 & +3.2{\tiny$\pm$1.2} & +2.0 & 0.164 \\
UR-FUNNY & 12.6 & 13.0 & 14.5 & +1.5{\tiny$\pm$0.5} & +0.9 & 0.096 \\
SSW60 & 70.3 & 71.0 & 72.3 & +1.3{\tiny$\pm$0.5} & +0.8 & 0.083 \\
AV-MNIST & 69.4 & 70.6 & 71.9 & +1.3{\tiny$\pm$0.4} & +0.8 & 0.051 \\
MOSI & 23.1 & 23.4 & 23.7 & +0.3{\tiny$\pm$0.9} & +0.3 & 0.034 \\
MOSEI & 8.9 & 9.2 & 9.4 & +0.2{\tiny$\pm$0.4} & +0.1 & 0.027 \\
VATEX & 70.4 & 71.9 & 72.4 & +0.5{\tiny$\pm$0.5} & +0.3 & 0.017 \\
MSR-VTT & 46.8 & 48.0 & 48.4 & +0.4{\tiny$\pm$0.4} & +0.3 & 0.012 \\
VB100 & 19.2 & 20.6 & 20.9 & +0.3{\tiny$\pm$0.5} & +0.3 & 0.009 \\
YouCook2 & 12.6 & 12.8 & 12.9 & +0.1{\tiny$\pm$0.3} & +0.0 & 0.006 \\
\bottomrule
\end{tabular}
\end{table}

\begin{table}[t]
\centering
\caption{BPI and $\Delta$BPI against the oracle $I^{(3)}$ on XOR-Mix for bridges of increasing capacity (nats, $K{=}64$). With $\rho{=}0$ no sample needs the triple.}
\label{tab:oracle}
\footnotesize
\setlength{\tabcolsep}{3.6pt}
\begin{tabular}{@{}l cc cc@{}}
\toprule
 & \multicolumn{2}{c}{$\rho{=}0$, $I^{(3)}{=}0$} & \multicolumn{2}{c}{$\rho{=}0.1$, $I^{(3)}{=}0.199$} \\
\cmidrule(lr){2-3}\cmidrule(l){4-5}
Bridge & BPI & $\Delta$BPI & BPI & $\Delta$BPI \\
\midrule
Random projections & 1.240 & 0.004 & 1.420 & 0.170 \\
Frozen encoders & 0.288 & 0.001 & 0.480 & 0.190 \\
\rowcolor{oursrow}Trained heads (default) & 0.011 & 0.001 & 0.205 & 0.193 \\
Pairwise MLP & 0.005 & 0.000 & 0.201 & 0.196 \\
\bottomrule
\end{tabular}
\end{table}

\begin{table}[t]
\centering
\caption{Left: Symile-M3-2 accuracy (\%) when only a share of the training tuples is complete. Right: Vision\&Touch top-1 accuracy (\%) with $M$ modalities.}
\label{tab:scaling}
\footnotesize
\setlength{\tabcolsep}{3.0pt}
\begin{tabular}{@{}l ccc ccc@{}}
\toprule
 & \multicolumn{3}{c}{Complete triples} & \multicolumn{3}{c}{Modalities} \\
\cmidrule(lr){2-4}\cmidrule(l){5-7}
Method & 100\% & 10\% & 1\% & $M{=}3$ & $M{=}4$ & $M{=}5$ \\
\midrule
Tri-CLIP & 47.5 & 47.2 & 46.9 & 41.2 & 43.0 & 43.8 \\
Symile & 93.8 & 89.0 & 47.9 & 44.6 & 40.3 & 36.9 \\
Gated Symile & 94.1 & \underline{89.5} & \underline{48.2} & 45.3 & 42.9 & 40.6 \\
H-CLIP & \underline{94.4} & 89.3 & 48.0 & \underline{45.8} & 42.7 & 39.5 \\
ConFu & 92.6 & 85.1 & 47.6 & 45.1 & \underline{46.2} & \underline{46.0} \\
\rowcolor{oursrow}\textbf{CHORD} & \textbf{94.8} & \textbf{92.4} & \textbf{66.0} & \textbf{46.9} & \textbf{47.8} & \textbf{48.6} \\
\bottomrule
\end{tabular}
\end{table}

\begin{figure}[t]
  \centering
  \includegraphics[width=\linewidth]{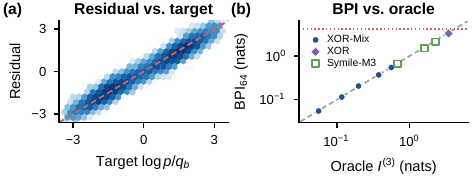}
  \vspace{-14pt}
  \caption{What the residual learns. (a) Residual versus its target $\log p/q_b$ under the trained bridge on synergy queries of XOR-Mix with a per-sample reliability, both centered per query. (b) $\BPI_{64}$ versus the oracle $I^{(3)}$. Hollow markers: $\log w$, an upper bound for Symile-M3. Dotted line: the $\log(K{+}1)$ cap.}
  \label{fig:bpi}
  \vspace{-8pt}
\end{figure}

The residual targets $\log p/q_b$, whereas the interaction structure of the data is $I^{(3)}=\KL(p\,\|\,q_2^\star)$, and the two differ by the misfit of the bridge. For \cref{fig:bpi}a we let the synergy rule of XOR-Mix hold only with a reliability drawn per sample, so that the target $\log p/q_b$ is finite and graded. The learned residual follows this target closely on synergy queries. In \cref{fig:bpi}b, BPI tracks the oracle $I^{(3)}$ on XOR and XOR-Mix, whose discrete factors give it exactly, and stays below $\log w$ on Symile-M3, which bounds the oracle of the recorded data from above. \Cref{tab:oracle} tests the decomposition directly. As the bridge weakens, its misfit inflates BPI, even on XOR-Mix with $\rho{=}0$, where no sample needs the triple and $I^{(3)}$ is zero, while $\Delta\BPI$ stays close to $I^{(3)}$ for every bridge. 
On real data the oracle is unknown. As shown in \cref{tab:decomp}, which orders the datasets by $\Delta\BPI$ and splits each gain into calibration and residual, $\Delta\BPI$ predicts where the residual helps. The residual adds most on MUStARD, UR-FUNNY, SSW60, and AV-MNIST, the datasets with the largest $\Delta\BPI$, and little elsewhere, where calibration contributes at least as much. Higher-order learning helps where the data hold interaction structure, not everywhere. The trained residual reads only the inputs, so it can also be added to the scores of the eight baselines without retraining them. The Plug-in column of \cref{tab:decomp} shows that these gains also follow $\Delta\BPI$, and a second seed of the same baseline or a pairwise residual trained on the same decoys adds less.

\subsection{Data Efficiency, More Modalities, and Cost}
\label{sec:scaling}

The results with missing modalities are shown in the left half of \cref{tab:scaling}, where each modality is masked independently so that only a share of training tuples is complete. The bridge needs only pairs and trains on every tuple with at least two observed modalities, while the residual trains on complete ones. With a tenth of the complete triples \method comes close to Symile trained on all of them, and with a hundredth it stays far above Symile. The right half adds modalities: the bridge conditions on all other modalities, and one residual sums over the triples that contain the target, which covers interactions up to order three. Symile and H-CLIP degrade as modalities are added, whereas \method improves. 
Sampling costs one matrix-vector product per query. Retrieval still combines the bridge and residual scores in one $d{+}R$-dimensional inner product per candidate.

\FloatBarrier
\section{Conclusion}
\label{sec:conclusion}

Higher-order multimodal objectives can represent joint structure yet rarely learn it, since a single pair rejects their negatives when pairs carry most of the evidence. In this work, we propose \method, which draws decoys from a calibrated pairwise model, trains a residual on what it misses, and separates connected information from the misfit of the model. The same separation tells where higher-order learning pays off, both for \method and for retrievers that receive its residual.




{
    \small
    \bibliographystyle{ieeenat_fullname}
    \bibliography{reference}
}
\clearpage
\setcounter{table}{0}
\setcounter{figure}{0}
\setcounter{equation}{0}
\setcounter{lemma}{0}
\renewcommand{\thetable}{S\arabic{table}}
\renewcommand{\thefigure}{S\arabic{figure}}
\renewcommand{\theequation}{S\arabic{equation}}
\renewcommand{\thelemma}{S\arabic{lemma}}
\renewcommand{\theHtable}{S\arabic{table}}
\renewcommand{\theHfigure}{S\arabic{figure}}
\renewcommand{\theHequation}{S\arabic{equation}}
\renewcommand{\theHlemma}{S\arabic{lemma}}
\appendix
\raggedbottom
\makeatletter
\setlength{\@fptop}{0pt}
\setlength{\@fpsep}{12pt}
\setlength{\@dblfptop}{0pt}
\setlength{\@dblfpsep}{12pt}
\makeatother

\section{Proofs and Conditions}
\label{supp:proofs}

\textbf{Proof of \cref{prop:starve}.} Let $\pi_n=e^{s_n}/\sum_{m=0}^{N-1}e^{s_m}$, with index $0$ for the positive. Since $\ell=-s_0+\log\sum_m e^{s_m}$ and $b$ does not depend on $\theta$,
\begin{equation}
\nabla_\theta\ell=-(1-\pi_0)\nabla_\theta r_0+\textstyle\sum_{n\ge1}\pi_n\nabla_\theta r_n ,
\end{equation}
so $\|\nabla_\theta\ell\|\le2G(1-\pi_0)=2G\sum_{n\ge1}\pi_n$. Each $\pi_n\le e^{s_n}/(e^{s_0}+e^{s_n})\le\min\{1,e^{s_n-s_0}\}$, and $s_n-s_0=-\Delta_n+r_n-r_0\le2B-\Delta_n$, which gives \cref{eq:starve}. For the expectation, bound each term by $1$ when $\Delta_n\le\delta$ and by $e^{2B-\delta}$ otherwise, and average over queries and over negatives drawn independently of $x_S$. Each term then has probability $\pi_\delta=P(\Delta_n\le\delta)$ of the first case. The bound covers only parameters on which $b$ does not depend. For parameters shared with the pairwise part, the gradient also contains the pairwise term. It bounds the gradient of one query and its average over queries, and does not describe the full optimization trajectory. \hfill$\square$

\begin{lemma}[Independent draws from the bank]
\label{lem:gumbel}
For logits $l_1,\dots,l_Q$ and i.i.d.\ $g_n\sim\mathrm{Gumbel}(0,1)$, $\arg\max_n(l_n+g_n)$ is distributed as $\mathrm{softmax}(l)$, and repeating with fresh noise gives $K$ independent draws~\cite{kool2019gumbel}. If the bank holds i.i.d.\ samples of $p(x_k)$, the bank softmax of \cref{eq:sampler} is a self-normalized importance sampler for $q_b(\cdot\mid x_S)$ with weights $e^{b_k}$, and expectations under it converge to those under $q_b$ as $Q\to\infty$.
\end{lemma}

\textbf{Proof of \cref{prop:consistency}.} (i) Fix $k$ and $x_S$, and write $p=p(\cdot\mid x_S)$ and $q=q_b(\cdot\mid x_S)$. Place the positive at a uniformly random index among $K{+}1$ candidates whose other entries are i.i.d.\ draws from $q$, independent of the positive. Given the candidates $y_0,\dots,y_K$, the posterior that index $j$ holds the positive is $w(y_j)/\sum_m w(y_m)$ with $w=p/q$. The decoy loss is the expected cross-entropy between this posterior and the softmax of $r$ over the candidates, so it is minimized exactly when $r-\log w$ does not depend on the candidate~\cite{ma2018nce}. The support condition holds because $q\propto p(x_k)e^{b_k}$ with finite $b_k$. With symmetric weights, $q_b(x_k\mid x_S)$ is the conditional of the joint $q_b(x)$, so $p(x_k\mid x_S)/q_b(x_k\mid x_S)=[p(x)/q_b(x)]\,[q_b(x_S)/p(x_S)]$ and $r(x)=\log p(x)/q_b(x)$ attains the optimum of every target up to a function of $x_S$. Without symmetry the three optima need not be compatible with one shared residual. (ii) By definition of $q_b$, $b_k=\log\big(q_b(x_k\mid x_S)/p(x_k)\big)+\log Z(x_S)$, and adding \cref{eq:rstar} gives $\log\big(p(x_k\mid x_S)/p(x_k)\big)$ plus a function of $x_S$. (iii) Let $P$ draw $y_0\sim p$ and $y_{1..K}\sim q$ independently, let $Q$ draw all $K{+}1$ candidates from $q$, and let $T=\log\big((K{+}1)e^{r(y_0)}/\sum_j e^{r(y_j)}\big)$. The Donsker-Varadhan inequality gives $\E_P[T]-\log\E_Q[e^T]\le\KL(P\|Q)=\KL(p\|q)$, and exchangeability under $Q$ gives $\E_Q[e^T]=1$. Since $\E_P[T]=\log(K{+}1)-\mathcal{L}_{\text{dec},k}$, averaging over $x_S$ and the three targets gives $\BPI_K\le\Ibp(q_b)$. Finally, $r^\star$ is constant in $x_k$ if and only if $p=q$ almost surely. These statements hold over all measurable $r$ and do not guarantee that the finite-rank head of \cref{eq:residual} reaches the optimum. \hfill$\square$

\textbf{Proof of \cref{prop:noharm}.} (i) One-to-one scores $\langle\vu_i,\vu_k\rangle$ depend only on $f_m$ and $g_m$. With frozen encoders, $f_m$ is fixed and $g_m$ receives no gradient from the decoy loss, so its trajectory is that of $\Lpair+\Lcal$ alone. With shared encoders the decoy loss reaches $f_m$ and changes $\vu_m=g_m(f_m(x_m))$, so no guarantee follows. (ii) If $p(\cdot\mid x_S)=q_b(\cdot\mid x_S)$, part (i) of \cref{prop:consistency} gives $r^\star=c(x_S)$. (iii) With $\vw=\mathbf{0}$, \cref{eq:residual} gives $r\equiv0$. \hfill$\square$

\subsection{What BPI Measures}
\label{supp:bpi}

\textbf{Proof of \cref{prop:decomp}.} Write $x_S$ for the pair of non-target modalities, $p_S$ and $q_S$ for their marginals, and $\mathcal{E}_2$ for the pairwise exponential family $\{q\propto\exp(\sum_m\theta_m(x_m)+\sum_{i<j}\theta_{ij}(x_i,x_j))\}$, which contains $q_b(x)\propto\prod_m p(x_m)\exp\sum_{i<j}\beta_{ij}\langle\vu_i,\vu_j\rangle/\tau_b$. By the chain rule, each target contributes $\E_{p_S}\KL(p(\cdot\mid x_S)\|q(\cdot\mid x_S))=\KL(p\|q)-\KL(p_S\|q_S)$, so $\Ibp(q)=\KL(p\|q)-\tfrac13\sum_k\KL(p_S\|q_S)$. The maximum-entropy model $q_2^\star$ with the pairwise marginals of $p$ is the m-projection of $p$ onto $\mathcal{E}_2$, and the Pythagorean relation of information geometry~\cite{amari2001hierarchy} gives $\KL(p\|q)=\KL(p\|q_2^\star)+\KL(q_2^\star\|q)$ for every $q\in\mathcal{E}_2$. Because $q_2^\star$ and $p$ share all pairwise marginals, $\KL(p_S\|q_S)=\KL(q_{2,S}^\star\|q_S)$. Substituting and applying the chain rule under $q_2^\star$ gives $\Ibp(q_b)=\KL(p\|q_2^\star)+\tfrac13\sum_k\E_{q_{2,S}^\star}\KL(q_2^\star(\cdot\mid x_S)\|q_b(\cdot\mid x_S))$. As in the statement, we assume that $q_2^\star$ lies in $\mathcal{E}_2$ and that all divergences are finite. Both hold for the discrete tasks and the Gaussian control used here. \hfill$\square$

\textbf{Oracle Values.} For each target, $\Ibp(q_2^\star)=\KL(p\|q_2^\star)=I^{(3)}$, so the oracle is the same for all three targets. On XOR all pairs are independent, $q_2^\star$ is the product of marginals, and $I^{(3)}=5\log2$. On XOR-Mix the code $z$ factors out, and the interaction between the flag $m$, the bit $e_1$, and the last target bit is invisible to pairs, so $q_2^\star$ gives the candidate $t=e_1$ probability $1-\rho/2$ on every query and $I^{(3)}=H_b(\rho/2)$, the binary entropy in nats. This is larger than $\rho\log2$ because the best pairwise model also hedges on samples that need no triple. These values also hold for the observed inputs. Each bit is observed with Gaussian noise of standard deviation 0.1, so its sign recovers it except with probability below $10^{-22}$. When each observation determines its discrete factor, the observed density is the discrete one times per-modality noise terms, which are unary terms of $\mathcal{E}_2$ and leave $I^{(3)}$ unchanged. On Symile-M3 the speech is pairwise independent of image and caption, so $q_2^\star$ makes it independent of both, and $I^{(3)}$ is the information between the speech and the other two modalities. For the discrete factors, class and language, this is $\log w$. The recorded images, speech, and captions pass these factors through per-modality channels that are not invertible, so by data processing $\log w$ bounds the $I^{(3)}$ of the recorded data from above, and \cref{fig:bpi}b marks these points as bounds. The Gaussian control has $I^{(3)}=0$. We compute the discrete values by iterative proportional fitting and compare them with BPI in \cref{fig:bpi}b and \cref{tab:oracle,tab:supp_bridgecap}.

\textbf{Misfit and $\Delta\BPI$.} If $x_3$ is a function of $x_1$ and $x_2$ is independent of both, then $I^{(3)}=0$, and a bridge that ignores $x_1$ has $M(q_b)=I(X_1;X_3)$. A pairwise residual trained on the same decoys extends the bridge within $\mathcal{E}_2$ and removes most of its misfit, so $\Delta\BPI=\BPI_K-\BPI_{\text{pair}}$ approximates $I^{(3)}$. It fits conditionals rather than the joint, and finite $K$, finite banks, and optimization error remain, so $\Delta\BPI$ is an estimate rather than a bound. Both readouts are capped at $\log(K{+}1)$, so their difference also shrinks when both approach this cap. On the real datasets BPI stays below 0.22 nats with $K{=}64$, far from $\log 65\approx4.17$. \Cref{tab:supp_bridgecap} tests it against the oracle across bridges.

\subsection{Drawing without Replacement}
\label{supp:sampling}

The posterior used in the proof of \cref{prop:consistency} requires negatives that are i.i.d.\ and independent of the positive. Drawing without replacement or removing the positive breaks this. Take two candidates with $p=q_b=(0.9,0.1)$, so the KL divergence is zero. If the positive is removed and the remaining candidate is the negative, the positive is identified by its own value, the optimal loss is the binary entropy $H(0.9)$, and $\log2-H(0.9)\approx0.368$ nats. The gap to i.i.d.\ draws is governed by the effective support $1/\sum_n q_b(n\mid x_S)^2$ rather than by $Q$, and \cref{tab:supp_sampling} reports both.

\subsection{The Pairwise Reference}
\label{supp:ref}

On synergy queries of XOR-Mix the two candidates share $z$ and differ in the last bit. An additive critic $f(x_1,x_3)+g(x_2,x_3)$ gives a score difference $F(e_1)+G(m,e_2)$. With $F(e_1)=e_1-\tfrac12$ and $G(1,e_2)=e_2$ it predicts $e_1\lor e_2$, which agrees with $e_1\oplus e_2$ on three of four inputs, and no additive rule does better because XOR is not linearly separable. Pairwise training fits each pair by likelihood, and on balanced XOR the likelihood-optimal additive score is constant, which gives 50\%. M3-Mix with $w{=}2$ has the same structure, since the speech language must match the language of one of two caption words. We therefore call 50\% the pairwise reference, not a ceiling for every pairwise critic.

\section{Benchmarks and Protocols}
\label{supp:bench}

\textbf{XOR.} We use the generator and protocol of Symile (App.~I.2 of~\cite{saporta2024symile}), which ConFu also uses: $a,b\in\{0,1\}^5$ are uniform, $i\sim\mathrm{Bern}(\hat p)$, and $c=a\oplus b$ if $i=1$ and $c=a$ otherwise, with 10K, 1K, and 5K tuples for training, validation, and test. Every method maps each modality with an affine encoder to $\mathbb{R}^{16}$ and retrieves $b$ from $(a,c)$ among its 32 values (chance 3.1\%). Because $I(a;b)=I(b;c)=0$ for every $\hat p$, only the joint dependence on $(a,c)$ helps, and the Bayes-optimal accuracy is $\hat p\cdot\tfrac{31}{32}+\tfrac1{32}$. \Cref{tab:controlled} uses $\hat p=1$.

\textbf{XOR-Mix.} The three modalities are $x_1=(z,e_1)$, $x_2=(m,e_2,n)$, and $x_3=(z,t)$, with a code $z\in\{0,1\}^{10}$, bits $e_1,e_2$, a flag $m$ with $P(m{=}1)=\rho$, and four nuisance bits $n$. The last target bit is $t=e_1\oplus e_2$ when $m=1$ and $t=e_1$ otherwise. Bits are mapped to $\pm1$ and observed with Gaussian noise of standard deviation 0.1. The pair $(x_1,x_3)$ shares $z$, so a pairwise model narrows any target to two candidates, and $(x_2,x_3)$ are independent. We train on $10^6$ tuples and test on $10^4$ queries with $m{=}1$, retrieving among all $2^{11}$ targets. Pairwise training reaches 50\% (\cref{supp:ref}).

\textbf{Symile-M3 and M3-Mix.} We use the released Symile-M3 generator and splits with $w\in\{2,5,10\}$~\cite{saporta2024symile}, in which the language of the speech selects which of the $w$ class names in the caption the image shows. M3-Mix changes only the caption: with probability $1-\rho$ it names a single class, the image class in the language of the speech, so text alone identifies the image. Accuracy is measured on test samples whose caption names two classes. For \cref{tab:scaling} (left) and \cref{fig:supp_eff}, each modality of a training tuple is masked independently with the probability that yields the stated share of complete tuples, as in Symile. A tuple can then keep three, two, one, or no modalities. The bridge trains on every observed pair: at 100\%, 30\%, 10\%, 3\%, and 1\% complete tuples, 100\%, 74\%, 45\%, 23\%, and 12\% of tuples keep at least two modalities, with 3.0, 1.34, 0.65, 0.29, and 0.14 observed pairs per tuple. The residual trains on complete tuples only.

\textbf{XOR-Mix with a Per-Sample Reliability.} For \cref{fig:bpi}a, $\rho=0.1$, $x_2$ also carries a reliability $c\sim\mathcal{U}[0,0.95]$, and for $m=1$ the last bit equals $e_1\oplus e_2$ with probability $(1{+}c)/2$, so on synergy queries the two candidates that share $z$ have $p(t\mid x_S)=(1\pm c)/2$. The residual targets $\log p/q_b$ under the trained bridge (\cref{prop:consistency}). On this discrete task we evaluate $q_b(\cdot\mid x_S)$ exactly by normalizing the bridge over all $2^{11}$ targets, compute the target for both candidates, and compare it with the learned residual after removing the per-query mean, which absorbs $c_k(x_S)$. The trained bridge gives the candidate $t=e_1$ a probability close to the $1-\rho/2$ of $q_2^\star$, so this target is close to $\log p/q_2^\star$, and the difference is the pointwise misfit of \cref{prop:decomp}.

\textbf{Critics and Negatives.} For \cref{tab:negatives}, each residual family replaces \cref{eq:residual} on top of the same calibrated bridge, with the same bank and $K{=}32$. Uniform proposals draw from the bank with logits $b+r$, decoys from \cref{eq:sampler} with logits $r$. Both are the noise-corrected logits $r+(1-\gamma)b$ at $\gamma=0$ and $\gamma=1$ and share the same target residual. The sweep of \cref{fig:sensitivity}b uses the same correction. Top-$K$ takes the $K$ bank targets with the highest bridge score, with logits $r$. It has no valid noise correction, and it differs from the limit $\gamma\to\infty$ of independent draws, which repeats the single highest-scoring target $K$ times. The pairwise control is $\sum_{i<j}\vw_{ij}^\top(\va_i\odot\va_j)$. For \cref{fig:diagnosis}b, the baselines use their own losses with in-batch negatives, and \method keeps $K{=}32$ decoys at every batch size.

\textbf{Synthetic-XNOR.} We use the generator of Gated Symile~\cite{rheude2026gated} with misalignment probability one, so one non-target modality of every sample carries the signal of another sample, and report top-1 accuracy among 128 candidates.

\textbf{Real Datasets.} MOSI, MOSEI, UR-FUNNY, and MUStARD use the MultiBench features, splits, and modality order of ConFu~\cite{liang2021multibench,koutoupis2026confu}. AV-MNIST pairs images and spoken digits with templated label text. Bird-MML pretraining and the zero-shot SSW60 and VB100 protocols follow ConFu. MSR-VTT uses the 1k-A test split, VATEX the public test videos that remain downloadable, and YouCook2 its validation clips. Audio is extracted from the videos, and the LanguageBind video, audio, and text encoders stay frozen. For text-to-(video, audio) retrieval, a candidate pair is ranked by $b_T+r$. With the shared optimal residual $\log p/q_b$, this score differs from the optimal score $\log p(x_T,x_V,x_A)/\big(p(x_T)\,p(x_V,x_A)\big)$ by $\log p(x_V,x_A)/\big(p(x_V)\,p(x_A)\big)-\beta_{VA}\langle\vu_V,\vu_A\rangle/\tau_b$ up to a constant, which is the misfit of the video-audio term of the bridge. Vision\&Touch~\cite{lee2019visiontouch} provides images, force, proprioception, depth, and contact signals. All datasets are publicly available.

\textbf{Decoy@K Probe.} For each test query, the judge bridge defines a distribution over all test targets except duplicates of the positive. We draw candidates with $\gamma=1$ and keep one only if each of its two pairwise scores under the judge reaches the median score of true pairs, redrawing otherwise, until 15 decoys are collected. Training decoys are drawn without this filter. The judge is a calibrated Tri-CLIP model trained with a different seed and initialization. For the LanguageBind datasets its heads are trained on a disjoint half of the training set. Rand@16 draws the 15 candidates uniformly instead (\cref{tab:supp_probe_sanity}). To test whether the probe favors models trained against the same family of bridge, we repeat it with a bilinear judge $\vu_i^\top W_{ik}\vu_k$ and with a pairwise MLP judge on concatenated embeddings, each calibrated in the same way and filtered with the same median rule, and \cref{tab:judges} reports the results.

\textbf{BPI and the Gaussian Control.} $\BPI_K$ uses held-out tuples, $K=64$ independent decoys from the model's own bridge, and the average over the three targets, with bootstrap intervals over queries. The Gaussian control draws $z=(z_1,z_2,z_3)$ from a zero-mean Gaussian with random pairwise cross-covariances, adds noise in this latent space, and maps each $z_m$ through a fixed invertible network of orthogonal layers and leaky ReLUs. Per-modality invertible maps add only per-modality Jacobian terms to $\log p(x)$, so the data have no interaction term and their interaction information is zero. Whether a given bridge represents the pairwise terms is a separate question. The weak-bridge variant replaces the trained heads by fixed random projections (\cref{tab:supp_bridgecap}).

\section{Implementation Details}
\label{supp:impl}

\begin{table}[t]
\centering
\caption{Settings per benchmark ($d$: embedding size, $R$: residual rank, $B$: batch size, ConFu: same as the ConFu configuration).}
\label{tab:supp_settings}
\footnotesize
\setlength{\tabcolsep}{2.0pt}
\begin{tabular}{@{}l l cccc@{}}
\toprule
Setting & Encoders & $d$ & $R$ & $B$ & Epochs \\
\midrule
XOR & affine & 16 & 16 & 1000 & 100 \\
XOR-Mix & linear & 16 & 64 & 256 & 50 \\
Symile-M3 & frozen (Symile) & 512 & 512 & 512 & 20 \\
MultiBench & ConFu & ConFu & 128 & ConFu & ConFu \\
Bird-MML & ConFu & ConFu & 256 & ConFu & ConFu \\
LanguageBind & frozen, MLP heads & 512 & 512 & 256 & 30 \\
\bottomrule
\end{tabular}
\end{table}

\begin{table}[t]
\centering
\caption{Cost on MSR-VTT with one A100 80GB, batch 256, and a FIFO bank. Dim.: size of the retrieval embedding. Time: median time per step over 200 steps after 50 warm-up steps, including bank sampling, relative to Tri-CLIP, with frozen and with end-to-end encoders. Mem.: extra peak memory. Lat.: time to score 10k precomputed candidates for one query, excluding data transfer.}
\label{tab:supp_cost}
\footnotesize
\setlength{\tabcolsep}{2.6pt}
\begin{tabular}{@{}l c cc cc@{}}
\toprule
 & & \multicolumn{2}{c}{Time} & & \\
\cmidrule(lr){3-4}
Method & Dim. & Frozen & E2E & Mem. (GB) & Lat. (ms) \\
\midrule
Tri-CLIP & 512 & 1.00 & 1.00 & 0.0 & 0.9 \\
Symile & 512 & 1.04 & 1.02 & 0.1 & 1.0 \\
Gated Symile & 512 & 1.31 & 1.18 & 0.6 & 6.8 \\
H-CLIP & 512 & 1.03 & 1.02 & 0.1 & 1.0 \\
GRAM & 512 & 1.06 & 1.03 & 0.1 & 2.7 \\
TRIANGLE & 512 & 1.05 & 1.03 & 0.1 & 2.4 \\
PMRL & 512 & 1.09 & 1.05 & 0.2 & 3.1 \\
ConFu & 512 & 1.12 & 1.06 & 0.3 & 1.2 \\
\rowcolor{oursrow}\textbf{CHORD} & 1024 & 1.07 & 1.04 & 0.4 & 1.6 \\
\method, $R{=}128$ & 640 & 1.05 & 1.03 & 0.2 & 1.1 \\
\bottomrule
\end{tabular}
\end{table}

\textbf{Statistics.} All main-paper numbers are means over five seeds. The main text reports no significance tests, so they are collected here. Against the strongest baseline, uncorrected paired $t$-tests over seeds give $p{<}0.05$ for \method in the following cases. In \cref{tab:controlled}, these are XOR-Mix and M3-Mix. In \cref{tab:multibench}, they are MOSEI two-to-one and Decoy@16 on MOSI, MOSEI, UR-FUNNY, and MUStARD. In \cref{tab:audiovisual}, they are MSR-VTT, VATEX, SSW60, and VB100, and in \cref{tab:scaling}, the settings with 10\% and 1\% complete triples and with four and five modalities. For the central claim, the gain of the residual over its own bridge, we test the ten real datasets with paired $t$-tests over seeds and apply Holm's correction~\cite{holm1979simple}. For the bootstrap intervals of \cref{tab:supp_decomp}, we first average each test query's metric over the five seeds and then resample queries with replacement, 10{,}000 times, and report percentile intervals. On MSR-VTT, VATEX, and YouCook2 we resample videos instead, so that all captions of a video stay together.

Symile-M3 uses frozen Whisper, CLIP, and XLM-RoBERTa encoders as in Symile~\cite{radford2023whisper,radford2021clip,conneau2020xlmr}. The memory bank is a first-in-first-out queue of detached target embeddings and features. For frozen encoders it is filled once from the training set. The residual uses LayerNorm on each projection, AdamW with weight decay $10^{-4}$, and the learning rate of the baseline it is compared with. \Cref{tab:supp_cost} reports the cost: sampling adds one matrix-vector product per query, and retrieval stays a single inner product per candidate.

\section{Additional Results}
\label{supp:results}

\begin{table*}[t]
\centering
\caption{Table~\ref{tab:controlled} with standard deviations over five seeds.}
\label{tab:supp_controlled_std}
\small
\setlength{\tabcolsep}{4.5pt}
\begin{tabular}{@{}l ccc cccc@{}}
\toprule
Method & XOR & XOR-Mix & Syn-XNOR & M3 ($w{=}2$) & M3 ($w{=}5$) & M3 ($w{=}10$) & M3-Mix \\
\midrule
Tri-CLIP & 3.8{\tiny$\pm$0.2} & 50.1{\tiny$\pm$0.4} & 24.4{\tiny$\pm$1.1} & 47.5{\tiny$\pm$0.1} & 18.8{\tiny$\pm$0.1} & 9.6{\tiny$\pm$0.1} & 48.6{\tiny$\pm$0.5} \\
Symile & \textbf{100.0{\tiny$\pm$0.0}} & 64.2{\tiny$\pm$3.1} & 33.2{\tiny$\pm$1.4} & 93.8{\tiny$\pm$0.1} & 92.0{\tiny$\pm$0.1} & 88.3{\tiny$\pm$0.1} & 66.2{\tiny$\pm$2.8} \\
Gated Symile & \underline{99.8{\tiny$\pm$0.1}} & \underline{66.0{\tiny$\pm$2.9}} & \textbf{87.1{\tiny$\pm$0.9}} & 94.1{\tiny$\pm$0.2} & 92.0{\tiny$\pm$0.2} & 88.6{\tiny$\pm$0.2} & 67.0{\tiny$\pm$2.6} \\
H-CLIP & \textbf{100.0{\tiny$\pm$0.0}} & 65.3{\tiny$\pm$3.3} & 41.6{\tiny$\pm$1.6} & \underline{94.4{\tiny$\pm$0.2}} & \underline{92.3{\tiny$\pm$0.2}} & \underline{88.9{\tiny$\pm$0.3}} & \underline{68.4{\tiny$\pm$2.9}} \\
GRAM & 13.8{\tiny$\pm$1.2} & 50.3{\tiny$\pm$0.5} & 48.3{\tiny$\pm$1.8} & 48.1{\tiny$\pm$0.4} & 19.3{\tiny$\pm$0.4} & 9.9{\tiny$\pm$0.3} & 48.9{\tiny$\pm$0.6} \\
TRIANGLE & 12.9{\tiny$\pm$1.1} & 50.6{\tiny$\pm$0.5} & 60.7{\tiny$\pm$1.5} & 48.4{\tiny$\pm$0.4} & 19.6{\tiny$\pm$0.3} & 10.1{\tiny$\pm$0.3} & 49.2{\tiny$\pm$0.6} \\
PMRL & 9.6{\tiny$\pm$0.9} & 50.2{\tiny$\pm$0.4} & 45.2{\tiny$\pm$1.7} & 47.8{\tiny$\pm$0.3} & 19.0{\tiny$\pm$0.3} & 9.7{\tiny$\pm$0.3} & 48.8{\tiny$\pm$0.5} \\
ConFu & \textbf{100.0{\tiny$\pm$0.0}} & 61.0{\tiny$\pm$3.4} & 52.7{\tiny$\pm$1.9} & 92.6{\tiny$\pm$0.3} & 89.8{\tiny$\pm$0.4} & 85.1{\tiny$\pm$0.4} & 63.1{\tiny$\pm$3.0} \\
\rowcolor{oursrow}\textbf{CHORD} & \textbf{100.0{\tiny$\pm$0.0}} & \textbf{97.9{\tiny$\pm$0.7}} & \underline{84.6{\tiny$\pm$1.0}} & \textbf{94.8{\tiny$\pm$0.2}} & \textbf{92.6{\tiny$\pm$0.2}} & \textbf{89.4{\tiny$\pm$0.3}} & \textbf{89.6{\tiny$\pm$1.2}} \\
\bottomrule
\end{tabular}
\end{table*}

\begin{table*}[t]
\centering
\caption{Per-direction Recall@10 (\%) on all four MultiBench datasets, mean $\pm$ std over five seeds.}
\label{tab:supp_multibench_full}
\scriptsize
\setlength{\tabcolsep}{3.0pt}
\begin{tabular}{@{}l ccc ccc ccc@{}}
\toprule
Target & \multicolumn{3}{c}{M1} & \multicolumn{3}{c}{M2} & \multicolumn{3}{c}{M3} \\
\cmidrule(lr){2-4}\cmidrule(lr){5-7}\cmidrule(l){8-10}
Query & M2 & M3 & M23 & M1 & M3 & M13 & M1 & M2 & M12 \\
\midrule
\multicolumn{10}{@{}l}{\textit{MOSI}} \\
Tri-CLIP & \textbf{23.3{\tiny$\pm$1.9}} & \textbf{20.7{\tiny$\pm$1.6}} & \underline{22.4{\tiny$\pm$1.8}} & \underline{23.2{\tiny$\pm$3.6}} & \underline{23.1{\tiny$\pm$1.8}} & \underline{23.6{\tiny$\pm$2.0}} & \textbf{21.1{\tiny$\pm$2.0}} & \underline{24.6{\tiny$\pm$2.1}} & \underline{23.4{\tiny$\pm$1.9}} \\
Symile & \textcolor{gray}{n/a} & \textcolor{gray}{n/a} & 15.9{\tiny$\pm$0.9} & \textcolor{gray}{n/a} & \textcolor{gray}{n/a} & 18.5{\tiny$\pm$2.4} & \textcolor{gray}{n/a} & \textcolor{gray}{n/a} & 16.7{\tiny$\pm$1.7} \\
Gated Symile & \textcolor{gray}{n/a} & \textcolor{gray}{n/a} & 17.6{\tiny$\pm$1.2} & \textcolor{gray}{n/a} & \textcolor{gray}{n/a} & 19.2{\tiny$\pm$2.1} & \textcolor{gray}{n/a} & \textcolor{gray}{n/a} & 18.4{\tiny$\pm$1.6} \\
H-CLIP & \textcolor{gray}{n/a} & \textcolor{gray}{n/a} & 17.4{\tiny$\pm$1.1} & \textcolor{gray}{n/a} & \textcolor{gray}{n/a} & 18.9{\tiny$\pm$2.2} & \textcolor{gray}{n/a} & \textcolor{gray}{n/a} & 18.0{\tiny$\pm$1.8} \\
GRAM & 5.8{\tiny$\pm$3.1} & 10.5{\tiny$\pm$2.9} & 16.0{\tiny$\pm$2.5} & 7.3{\tiny$\pm$2.3} & 0.2{\tiny$\pm$0.6} & 7.6{\tiny$\pm$3.4} & 12.0{\tiny$\pm$4.1} & 0.3{\tiny$\pm$0.2} & 11.6{\tiny$\pm$4.2} \\
TRIANGLE & \textcolor{gray}{n/a} & \textcolor{gray}{n/a} & 8.5{\tiny$\pm$1.3} & \textcolor{gray}{n/a} & \textcolor{gray}{n/a} & 4.8{\tiny$\pm$0.7} & \textcolor{gray}{n/a} & \textcolor{gray}{n/a} & 9.2{\tiny$\pm$1.4} \\
PMRL & 19.4{\tiny$\pm$2.0} & 16.2{\tiny$\pm$1.8} & 21.0{\tiny$\pm$1.9} & 18.9{\tiny$\pm$2.6} & 17.5{\tiny$\pm$1.9} & 22.1{\tiny$\pm$2.2} & 17.1{\tiny$\pm$1.8} & 18.3{\tiny$\pm$2.3} & 21.7{\tiny$\pm$2.0} \\
ConFu & 21.2{\tiny$\pm$1.7} & 16.8{\tiny$\pm$2.3} & 17.1{\tiny$\pm$1.8} & 19.5{\tiny$\pm$1.4} & 20.8{\tiny$\pm$2.4} & 21.4{\tiny$\pm$1.9} & 15.8{\tiny$\pm$1.7} & 23.4{\tiny$\pm$2.7} & 20.9{\tiny$\pm$2.9} \\
\rowcolor{oursrow}\textbf{CHORD} & \underline{23.1{\tiny$\pm$1.6}} & \underline{20.6{\tiny$\pm$1.5}} & \textbf{23.2{\tiny$\pm$1.7}} & \textbf{23.8{\tiny$\pm$2.9}} & \textbf{23.5{\tiny$\pm$1.7}} & \textbf{23.9{\tiny$\pm$1.9}} & \underline{21.0{\tiny$\pm$1.8}} & \textbf{24.8{\tiny$\pm$2.0}} & \textbf{24.0{\tiny$\pm$1.8}} \\
\midrule
\multicolumn{10}{@{}l}{\textit{MOSEI}} \\
Tri-CLIP & \underline{7.4{\tiny$\pm$0.5}} & \underline{5.5{\tiny$\pm$0.6}} & \underline{8.6{\tiny$\pm$0.3}} & \textbf{7.6{\tiny$\pm$0.5}} & \textbf{7.8{\tiny$\pm$0.4}} & \underline{9.1{\tiny$\pm$0.5}} & 6.1{\tiny$\pm$0.3} & 6.5{\tiny$\pm$0.5} & \underline{9.0{\tiny$\pm$0.5}} \\
Symile & \textcolor{gray}{n/a} & \textcolor{gray}{n/a} & 6.5{\tiny$\pm$0.5} & \textcolor{gray}{n/a} & \textcolor{gray}{n/a} & 7.2{\tiny$\pm$0.7} & \textcolor{gray}{n/a} & \textcolor{gray}{n/a} & 7.7{\tiny$\pm$0.5} \\
Gated Symile & \textcolor{gray}{n/a} & \textcolor{gray}{n/a} & 7.0{\tiny$\pm$0.7} & \textcolor{gray}{n/a} & \textcolor{gray}{n/a} & 7.5{\tiny$\pm$0.9} & \textcolor{gray}{n/a} & \textcolor{gray}{n/a} & 8.3{\tiny$\pm$0.7} \\
H-CLIP & \textcolor{gray}{n/a} & \textcolor{gray}{n/a} & 6.8{\tiny$\pm$0.6} & \textcolor{gray}{n/a} & \textcolor{gray}{n/a} & 7.6{\tiny$\pm$0.5} & \textcolor{gray}{n/a} & \textcolor{gray}{n/a} & 7.8{\tiny$\pm$0.7} \\
GRAM & 3.3{\tiny$\pm$0.6} & 2.3{\tiny$\pm$0.6} & 4.6{\tiny$\pm$0.6} & 3.0{\tiny$\pm$0.7} & 3.1{\tiny$\pm$0.7} & 5.5{\tiny$\pm$0.7} & 2.9{\tiny$\pm$0.9} & 2.8{\tiny$\pm$0.8} & 5.4{\tiny$\pm$0.5} \\
TRIANGLE & \textcolor{gray}{n/a} & \textcolor{gray}{n/a} & 5.6{\tiny$\pm$0.5} & \textcolor{gray}{n/a} & \textcolor{gray}{n/a} & 6.0{\tiny$\pm$0.5} & \textcolor{gray}{n/a} & \textcolor{gray}{n/a} & 5.9{\tiny$\pm$0.6} \\
PMRL & 7.1{\tiny$\pm$0.5} & 4.5{\tiny$\pm$0.5} & 7.1{\tiny$\pm$0.5} & 6.9{\tiny$\pm$0.4} & 6.3{\tiny$\pm$0.5} & 8.5{\tiny$\pm$0.6} & 4.5{\tiny$\pm$0.6} & 5.9{\tiny$\pm$0.4} & 7.9{\tiny$\pm$0.4} \\
ConFu & 7.1{\tiny$\pm$0.5} & 5.1{\tiny$\pm$0.7} & 8.1{\tiny$\pm$0.8} & 6.6{\tiny$\pm$0.4} & 6.1{\tiny$\pm$0.4} & 8.6{\tiny$\pm$0.8} & \underline{6.2{\tiny$\pm$0.4}} & \underline{6.7{\tiny$\pm$0.4}} & 8.6{\tiny$\pm$0.5} \\
\rowcolor{oursrow}\textbf{CHORD} & \textbf{8.5{\tiny$\pm$0.6}} & \textbf{6.0{\tiny$\pm$0.6}} & \textbf{9.1{\tiny$\pm$0.5}} & \underline{7.0{\tiny$\pm$0.4}} & \underline{7.0{\tiny$\pm$0.5}} & \textbf{9.4{\tiny$\pm$0.4}} & \textbf{6.7{\tiny$\pm$0.4}} & \textbf{6.9{\tiny$\pm$0.3}} & \textbf{9.6{\tiny$\pm$0.4}} \\
\midrule
\multicolumn{10}{@{}l}{\textit{UR-FUNNY}} \\
Tri-CLIP & \underline{3.6{\tiny$\pm$0.6}} & \underline{3.9{\tiny$\pm$0.6}} & \underline{4.2{\tiny$\pm$0.5}} & \underline{3.7{\tiny$\pm$0.5}} & \textbf{16.4{\tiny$\pm$1.6}} & 16.5{\tiny$\pm$1.2} & 3.9{\tiny$\pm$0.4} & \underline{16.6{\tiny$\pm$1.2}} & 17.1{\tiny$\pm$1.1} \\
Symile & \textcolor{gray}{n/a} & \textcolor{gray}{n/a} & 3.6{\tiny$\pm$0.5} & \textcolor{gray}{n/a} & \textcolor{gray}{n/a} & 15.7{\tiny$\pm$0.9} & \textcolor{gray}{n/a} & \textcolor{gray}{n/a} & 16.1{\tiny$\pm$0.6} \\
Gated Symile & \textcolor{gray}{n/a} & \textcolor{gray}{n/a} & 3.9{\tiny$\pm$0.4} & \textcolor{gray}{n/a} & \textcolor{gray}{n/a} & 15.8{\tiny$\pm$0.9} & \textcolor{gray}{n/a} & \textcolor{gray}{n/a} & 17.2{\tiny$\pm$0.8} \\
H-CLIP & \textcolor{gray}{n/a} & \textcolor{gray}{n/a} & 4.0{\tiny$\pm$0.5} & \textcolor{gray}{n/a} & \textcolor{gray}{n/a} & 15.5{\tiny$\pm$1.0} & \textcolor{gray}{n/a} & \textcolor{gray}{n/a} & 16.8{\tiny$\pm$0.9} \\
GRAM & 3.4{\tiny$\pm$0.5} & 3.1{\tiny$\pm$0.9} & 4.1{\tiny$\pm$0.8} & 3.4{\tiny$\pm$0.9} & 0.2{\tiny$\pm$0.2} & 3.2{\tiny$\pm$0.4} & \underline{4.1{\tiny$\pm$0.4}} & 0.2{\tiny$\pm$0.0} & 3.5{\tiny$\pm$0.5} \\
TRIANGLE & \textcolor{gray}{n/a} & \textcolor{gray}{n/a} & 4.1{\tiny$\pm$0.4} & \textcolor{gray}{n/a} & \textcolor{gray}{n/a} & 3.2{\tiny$\pm$0.7} & \textcolor{gray}{n/a} & \textcolor{gray}{n/a} & 3.4{\tiny$\pm$0.9} \\
PMRL & 3.3{\tiny$\pm$0.5} & 3.6{\tiny$\pm$0.5} & 3.8{\tiny$\pm$0.6} & 3.4{\tiny$\pm$0.5} & 14.2{\tiny$\pm$1.3} & 13.5{\tiny$\pm$1.1} & 3.5{\tiny$\pm$0.5} & 13.4{\tiny$\pm$1.2} & 15.1{\tiny$\pm$1.0} \\
ConFu & 3.3{\tiny$\pm$0.3} & 3.7{\tiny$\pm$0.3} & 3.8{\tiny$\pm$0.3} & 3.4{\tiny$\pm$0.2} & \underline{14.8{\tiny$\pm$0.8}} & \underline{16.6{\tiny$\pm$0.9}} & 3.7{\tiny$\pm$0.4} & 15.8{\tiny$\pm$0.8} & \underline{20.0{\tiny$\pm$1.1}} \\
\rowcolor{oursrow}\textbf{CHORD} & \textbf{3.8{\tiny$\pm$0.4}} & \textbf{4.1{\tiny$\pm$0.4}} & \textbf{4.3{\tiny$\pm$0.4}} & \textbf{3.9{\tiny$\pm$0.4}} & \textbf{16.4{\tiny$\pm$1.0}} & \textbf{17.6{\tiny$\pm$0.9}} & \textbf{4.2{\tiny$\pm$0.4}} & \textbf{17.4{\tiny$\pm$0.9}} & \textbf{21.5{\tiny$\pm$1.0}} \\
\midrule
\multicolumn{10}{@{}l}{\textit{MUStARD}} \\
Tri-CLIP & 70.5{\tiny$\pm$4.1} & 28.7{\tiny$\pm$3.6} & 72.1{\tiny$\pm$3.9} & 70.9{\tiny$\pm$4.3} & 26.2{\tiny$\pm$3.2} & 71.6{\tiny$\pm$4.0} & 30.0{\tiny$\pm$3.4} & 24.4{\tiny$\pm$2.8} & 33.3{\tiny$\pm$3.3} \\
Symile & \textcolor{gray}{n/a} & \textcolor{gray}{n/a} & 61.1{\tiny$\pm$6.3} & \textcolor{gray}{n/a} & \textcolor{gray}{n/a} & 56.9{\tiny$\pm$3.0} & \textcolor{gray}{n/a} & \textcolor{gray}{n/a} & 21.6{\tiny$\pm$4.4} \\
Gated Symile & \textcolor{gray}{n/a} & \textcolor{gray}{n/a} & 63.9{\tiny$\pm$5.1} & \textcolor{gray}{n/a} & \textcolor{gray}{n/a} & 60.2{\tiny$\pm$3.4} & \textcolor{gray}{n/a} & \textcolor{gray}{n/a} & 23.8{\tiny$\pm$4.0} \\
H-CLIP & \textcolor{gray}{n/a} & \textcolor{gray}{n/a} & 62.8{\tiny$\pm$5.5} & \textcolor{gray}{n/a} & \textcolor{gray}{n/a} & 58.9{\tiny$\pm$3.6} & \textcolor{gray}{n/a} & \textcolor{gray}{n/a} & 22.6{\tiny$\pm$4.1} \\
GRAM & 61.5{\tiny$\pm$8.0} & 24.5{\tiny$\pm$5.3} & \underline{81.4{\tiny$\pm$4.1}} & 66.9{\tiny$\pm$5.4} & 25.0{\tiny$\pm$5.3} & 67.6{\tiny$\pm$5.4} & 27.7{\tiny$\pm$8.7} & 5.0{\tiny$\pm$2.1} & 31.5{\tiny$\pm$9.2} \\
TRIANGLE & \textcolor{gray}{n/a} & \textcolor{gray}{n/a} & 69.1{\tiny$\pm$4.2} & \textcolor{gray}{n/a} & \textcolor{gray}{n/a} & 59.7{\tiny$\pm$3.7} & \textcolor{gray}{n/a} & \textcolor{gray}{n/a} & 22.1{\tiny$\pm$5.0} \\
PMRL & 66.9{\tiny$\pm$4.6} & 27.3{\tiny$\pm$3.9} & 76.2{\tiny$\pm$4.2} & 67.4{\tiny$\pm$4.8} & 25.0{\tiny$\pm$3.6} & 71.4{\tiny$\pm$4.4} & 28.6{\tiny$\pm$3.8} & 22.4{\tiny$\pm$3.1} & 23.7{\tiny$\pm$3.9} \\
ConFu & \textbf{73.6{\tiny$\pm$3.2}} & \textbf{34.0{\tiny$\pm$3.1}} & 79.3{\tiny$\pm$3.6} & \textbf{74.7{\tiny$\pm$2.7}} & \underline{27.9{\tiny$\pm$3.2}} & \underline{75.0{\tiny$\pm$4.1}} & \textbf{34.2{\tiny$\pm$4.2}} & \textbf{28.2{\tiny$\pm$2.9}} & \underline{33.8{\tiny$\pm$3.2}} \\
\rowcolor{oursrow}\textbf{CHORD} & \underline{72.4{\tiny$\pm$3.5}} & \underline{30.8{\tiny$\pm$3.0}} & \textbf{81.6{\tiny$\pm$3.1}} & \underline{71.9{\tiny$\pm$3.8}} & \textbf{28.2{\tiny$\pm$2.9}} & \textbf{76.3{\tiny$\pm$3.3}} & \underline{33.1{\tiny$\pm$3.6}} & \underline{25.2{\tiny$\pm$2.6}} & \textbf{35.3{\tiny$\pm$3.0}} \\
\bottomrule
\end{tabular}
\end{table*}

\begin{table*}[t]
\centering
\caption{Zero-shot accuracy (\%) with audio only (A), vision only (V), and fused inputs (A+V).}
\label{tab:supp_zeroshot}
\small
\setlength{\tabcolsep}{5.5pt}
\begin{tabular}{@{}l ccc ccc ccc@{}}
\toprule
 & \multicolumn{3}{c}{SSW60} & \multicolumn{3}{c}{VB100} & \multicolumn{3}{c}{AV-MNIST} \\
\cmidrule(lr){2-4}\cmidrule(lr){5-7}\cmidrule(l){8-10}
Method & A & V & A+V & A & V & A+V & A & V & A+V \\
\midrule
Tri-CLIP & \underline{30.9} & 68.9 & 70.3 & \underline{3.7} & \underline{20.5} & \underline{19.2} & 39.8 & 62.6 & 69.4 \\
Symile & \textcolor{gray}{n/a} & \textcolor{gray}{n/a} & 60.4 & \textcolor{gray}{n/a} & \textcolor{gray}{n/a} & 13.1 & \textcolor{gray}{n/a} & \textcolor{gray}{n/a} & 71.0 \\
Gated Symile & \textcolor{gray}{n/a} & \textcolor{gray}{n/a} & 63.1 & \textcolor{gray}{n/a} & \textcolor{gray}{n/a} & 17.2 & \textcolor{gray}{n/a} & \textcolor{gray}{n/a} & \underline{71.5} \\
H-CLIP & \textcolor{gray}{n/a} & \textcolor{gray}{n/a} & 61.0 & \textcolor{gray}{n/a} & \textcolor{gray}{n/a} & 13.9 & \textcolor{gray}{n/a} & \textcolor{gray}{n/a} & 71.3 \\
GRAM & 0.8 & 66.5 & 57.1 & 1.2 & 13.5 & 7.9 & 9.5 & \underline{64.1} & 64.1 \\
TRIANGLE & \textcolor{gray}{n/a} & \textcolor{gray}{n/a} & 64.2 & \textcolor{gray}{n/a} & \textcolor{gray}{n/a} & 11.8 & \textcolor{gray}{n/a} & \textcolor{gray}{n/a} & 65.1 \\
PMRL & 27.4 & 67.8 & 66.2 & 3.1 & 18.9 & 15.1 & 37.6 & 62.9 & 66.0 \\
ConFu & 30.0 & \textbf{69.7} & \underline{71.2} & 3.2 & 19.5 & 18.3 & \underline{39.9} & \textbf{64.7} & 70.9 \\
\rowcolor{oursrow}\textbf{CHORD} & \textbf{31.3} & \underline{69.2} & \textbf{72.3} & \textbf{3.9} & \textbf{20.7} & \textbf{20.9} & \textbf{40.2} & 62.8 & \textbf{71.9} \\
\bottomrule
\end{tabular}
\end{table*}

\begin{table*}[t]
\centering
\caption{Video retrieval with frozen LanguageBind encoders, Recall@$k$ (\%). MSR-VTT: 1k-A test split, one caption per video, 1,000 candidates. VATEX: 1,500 test videos, one caption per video.}
\label{tab:supp_video}
\small
\setlength{\tabcolsep}{3.6pt}
\begin{tabular}{@{}l ccc ccc ccc ccc@{}}
\toprule
 & \multicolumn{6}{c}{MSR-VTT} & \multicolumn{6}{c}{VATEX} \\
\cmidrule(lr){2-7}\cmidrule(l){8-13}
 & \multicolumn{3}{c}{T$\to$VA} & \multicolumn{3}{c}{VA$\to$T} & \multicolumn{3}{c}{T$\to$VA} & \multicolumn{3}{c}{VA$\to$T} \\
\cmidrule(lr){2-4}\cmidrule(lr){5-7}\cmidrule(lr){8-10}\cmidrule(l){11-13}
Method & R@1 & R@5 & R@10 & R@1 & R@5 & R@10 & R@1 & R@5 & R@10 & R@1 & R@5 & R@10 \\
\midrule
Tri-CLIP & 46.8 & 72.1 & 81.3 & 44.0 & 70.6 & 80.2 & 70.4 & 93.0 & 96.8 & 66.9 & 91.7 & 96.0 \\
Symile & 44.9 & 70.4 & 79.8 & 42.6 & 69.1 & 78.9 & 67.9 & 91.6 & 95.9 & 64.8 & 90.3 & 95.2 \\
Gated Symile & 45.6 & 71.0 & 80.4 & 43.1 & 69.8 & 79.5 & 68.8 & 92.1 & 96.2 & 65.5 & 90.9 & 95.5 \\
H-CLIP & 45.1 & 70.6 & 80.0 & 42.8 & 69.3 & 79.1 & 68.1 & 91.8 & 96.0 & 65.0 & 90.5 & 95.3 \\
GRAM & 47.3 & 72.6 & 81.8 & 44.9 & 71.2 & 80.9 & 71.2 & 93.4 & 97.0 & 67.8 & 92.2 & 96.3 \\
TRIANGLE & \underline{47.8} & \underline{73.0} & \underline{82.0} & \underline{45.2} & \underline{71.6} & \underline{81.1} & \underline{71.8} & \underline{93.6} & \underline{97.1} & \underline{68.3} & \underline{92.4} & \underline{96.5} \\
PMRL & 47.1 & 72.4 & 81.6 & 44.6 & 71.0 & 80.6 & 71.0 & 93.3 & 96.9 & 67.6 & 92.0 & 96.2 \\
ConFu & 47.2 & 72.5 & 81.5 & 44.8 & 71.1 & 80.8 & 71.3 & 93.3 & 97.0 & 67.9 & 92.1 & 96.3 \\
\rowcolor{oursrow}\textbf{CHORD} & \textbf{48.4} & \textbf{73.4} & \textbf{82.5} & \textbf{45.8} & \textbf{72.0} & \textbf{81.6} & \textbf{72.4} & \textbf{93.9} & \textbf{97.3} & \textbf{68.9} & \textbf{92.7} & \textbf{96.7} \\
\bottomrule
\end{tabular}
\end{table*}

\textbf{Full Tables.} \Cref{tab:supp_controlled_std,tab:supp_multibench_full} report standard deviations and per-direction retrieval. In 11 of the 12 target directions, the two-to-one score of \method exceeds both of its single-pair scores. The exception is M3 on MOSI, where M3 from M2 alone is 0.8\% higher. \method gains most on the hardest MUStARD target, M3 from M12 (35.3 vs.\ 33.8 for ConFu), whereas ConFu falls below its own single pairs on MOSI, as its authors report~\cite{koutoupis2026confu}. \Cref{tab:supp_zeroshot} adds unimodal zero-shot accuracy. The audio and vision columns of \method equal those of its bridge, and the fused gain therefore comes from calibration and the residual. \Cref{tab:supp_video} reports all recall levels and both directions for video retrieval.

\begin{table}[t]
\centering
\caption{Linear-probe classification accuracy (\%).}
\label{tab:supp_probe}
\footnotesize
\setlength{\tabcolsep}{4pt}
\begin{tabular}{@{}l cccc@{}}
\toprule
Method & MOSI & UR-FUNNY & MUStARD & AV-MNIST \\
\midrule
Tri-CLIP & 63.2 & 64.2 & 61.8 & 71.8 \\
Symile & 67.3 & 64.5 & 60.6 & 73.1 \\
Gated Symile & \underline{67.8} & 64.9 & 61.9 & 73.4 \\
H-CLIP & 67.6 & 64.8 & 61.2 & 73.2 \\
GRAM & 65.4 & 64.5 & 64.4 & 70.6 \\
TRIANGLE & 65.2 & 64.6 & 64.3 & 70.9 \\
PMRL & 65.1 & 64.6 & 63.8 & 71.0 \\
ConFu & 66.5 & 65.1 & 64.3 & 73.6 \\
CoMM & 67.1 & 65.2 & 64.0 & 73.9 \\
InfMasking & \textbf{67.9} & \textbf{65.6} & \underline{64.5} & \textbf{74.2} \\
OVA-IB & 66.4 & 65.0 & 63.9 & 73.0 \\
\rowcolor{oursrow}\textbf{CHORD} & 67.4 & \underline{65.3} & \textbf{64.9} & \underline{74.0} \\
\bottomrule
\end{tabular}
\end{table}

\begin{table}[t]
\centering
\caption{Details of \cref{tab:decomp}. Cal.: bridge minus sum. Res.: \method minus bridge with its bootstrap 95\% interval over test queries. $^\ddagger$: $p{<}0.05$ in a paired $t$-test over seeds after Holm correction over the ten datasets. BPI and BPI$_{\text{pair}}$: decoy-loss bounds of the product residual and of a pairwise residual on the same decoys (nats, $K{=}64$).}
\label{tab:supp_decomp}
\footnotesize
\setlength{\tabcolsep}{2.6pt}
\begin{tabular}{@{}l cc c ccc@{}}
\toprule
Dataset & Cal. & Res. & 95\% CI & BPI & BPI$_{\text{pair}}$ & $\Delta$BPI \\
\midrule
MUStARD & +2.2 & +3.2$^\ddagger$ & [1.1, 5.4] & 0.214 & 0.050 & 0.164 \\
UR-FUNNY & +0.4 & +1.5$^\ddagger$ & [0.6, 2.4] & 0.136 & 0.040 & 0.096 \\
SSW60 & +0.7 & +1.3$^\ddagger$ & [0.7, 1.9] & 0.118 & 0.035 & 0.083 \\
AV-MNIST & +1.2 & +1.3$^\ddagger$ & [0.8, 1.8] & 0.081 & 0.030 & 0.051 \\
MOSI & +0.3 & +0.3 & [-0.6, 1.2] & 0.062 & 0.028 & 0.034 \\
MOSEI & +0.3 & +0.2 & [-0.1, 0.5] & 0.058 & 0.031 & 0.027 \\
VATEX & +1.5 & +0.5 & [-0.1, 1.1] & 0.041 & 0.024 & 0.017 \\
MSR-VTT & +1.2 & +0.4 & [-0.4, 1.2] & 0.034 & 0.022 & 0.012 \\
VB100 & +1.4 & +0.3 & [-0.3, 0.9] & 0.029 & 0.020 & 0.009 \\
YouCook2 & +0.2 & +0.1 & [-0.3, 0.5] & 0.021 & 0.015 & 0.006 \\
\bottomrule
\end{tabular}
\end{table}

\begin{table}[t]
\centering
\caption{Decoy sampling schemes, applied in training and in the BPI readout. BPI uses the default trained bridge and $K{=}64$ as in \cref{tab:supp_bridgecap}. The oracle $I^{(3)}$ is 0 on the Gaussian control and 0.199 nats on XOR-Mix with $\rho{=}0.1$. Bottom: median effective support $1/\sum_n q_b(n\mid x_S)^2$ of the bank distribution.}
\label{tab:supp_sampling}
\footnotesize
\setlength{\tabcolsep}{2.4pt}
\begin{tabular}{@{}l cc cc@{}}
\toprule
 & \multicolumn{2}{c}{BPI (nats)} & \multicolumn{2}{c}{Accuracy} \\
\cmidrule(lr){2-3}\cmidrule(l){4-5}
Draws & Gauss. & XOR-Mix & M3-Mix & MUStARD \\
\midrule
Independent (CHORD) & 0.002 & 0.205 & 89.6 & 64.4 \\
No replacement & 0.004 & 1.86 & 89.4 & 64.3 \\
No repl., positive removed & 0.006 & 3.95 & 89.1 & 64.0 \\
\bottomrule
\end{tabular}

\vspace{4pt}
\begin{tabular}{@{}cccccc@{}}
Gauss. & XOR-Mix & M3-Mix & MUStARD & SSW60 & MSR-VTT \\
\midrule
640 & 9 & 27 & 46 & 64 & 182 \\
\bottomrule
\end{tabular}
\end{table}

\begin{table*}[t]
\centering
\caption{BPI depends on the bridge, while $\Delta$BPI depends on it much less. Bridges of increasing capacity on XOR-Mix with $\rho{=}0$ (oracle $I^{(3)}=0$) and $\rho{=}0.1$ (oracle 0.199), the Gaussian control (oracle 0), and MUStARD (no oracle), in nats with $K{=}64$. The XOR-Mix blocks repeat \cref{tab:oracle} with $\BPI_{\text{pair}}$.}
\label{tab:supp_bridgecap}
\small
\setlength{\tabcolsep}{3.2pt}
\begin{tabular}{@{}l ccc ccc ccc ccc@{}}
\toprule
 & \multicolumn{3}{c}{XOR-Mix, $\rho{=}0$} & \multicolumn{3}{c}{XOR-Mix, $\rho{=}0.1$} & \multicolumn{3}{c}{Gaussian control} & \multicolumn{3}{c}{MUStARD} \\
\cmidrule(lr){2-4}\cmidrule(lr){5-7}\cmidrule(lr){8-10}\cmidrule(l){11-13}
Bridge & BPI & BPI$_{\text{pair}}$ & $\Delta$BPI & BPI & BPI$_{\text{pair}}$ & $\Delta$BPI & BPI & BPI$_{\text{pair}}$ & $\Delta$BPI & BPI & BPI$_{\text{pair}}$ & $\Delta$BPI \\
\midrule
Random frozen projections & 1.240 & 1.236 & 0.004 & 1.420 & 1.250 & 0.170 & 0.214 & 0.210 & 0.004 & 0.395 & 0.226 & 0.169 \\
Frozen encoders, calibrated & 0.288 & 0.287 & 0.001 & 0.480 & 0.290 & 0.190 & 0.031 & 0.029 & 0.002 & 0.262 & 0.101 & 0.161 \\
\rowcolor{oursrow}Trained heads, calibrated (default) & 0.011 & 0.010 & 0.001 & 0.205 & 0.012 & 0.193 & 0.002 & 0.002 & 0.000 & 0.214 & 0.050 & 0.164 \\
Pairwise MLP critic & 0.005 & 0.005 & 0.000 & 0.201 & 0.005 & 0.196 & 0.001 & 0.001 & 0.000 & 0.181 & 0.019 & 0.162 \\
\bottomrule
\end{tabular}
\end{table*}

\begin{table}[t]
\centering
\caption{Valid alternatives among the negatives. Suppression: mean $r(\text{positive})-r(\text{valid alternative})$ in nats (lower is better), for same-class images on M3-Mix and other captions of the same video on MSR-VTT. R@1: MSR-VTT VA$\to$T with all 20 captions of the 1,000 test videos as candidates, where any caption of the query video counts as correct, so it is higher than the one-caption protocol of \cref{tab:supp_video}.}
\label{tab:supp_falseneg}
\footnotesize
\setlength{\tabcolsep}{3pt}
\begin{tabular}{@{}l cc c@{}}
\toprule
 & \multicolumn{2}{c}{Suppression} & \\
\cmidrule(lr){2-3}
Proposal & M3-Mix & MSR-VTT & R@1 \\
\midrule
Uniform (bank) & 0.04 & 0.07 & 72.4 \\
Top-$K$ hardest & 1.21 & 0.86 & 69.8 \\
Hardness reweighting & 0.48 & 0.39 & 71.0 \\
\rowcolor{oursrow}Decoys from the bridge & 0.09 & 0.12 & 74.1 \\
\bottomrule
\end{tabular}
\end{table}

\textbf{Linear Probes.} \Cref{tab:supp_probe} trains linear classifiers on frozen embeddings. \method is competitive but not uniformly best: InfMasking leads on three of the four datasets. Probes reward linearly decodable labels rather than beyond-pair retrieval~\cite[cf.][]{zhang2026brainriem}, consistent with earlier observations that expressive objectives change probe accuracy little~\cite{rheude2026gpe}.

\begin{figure*}[t]
  \centering
  \includegraphics[width=\textwidth]{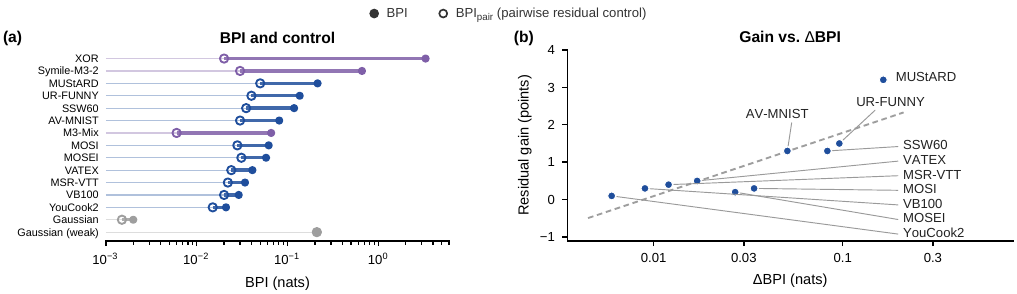}
  \caption{(a) BPI (filled) and $\BPI_{\text{pair}}$ of a pairwise residual on the same decoys (ring), whose gap is $\Delta\BPI$. (b) Gain of the residual over the calibrated bridge versus $\Delta\BPI$ on the ten real datasets.}
  \label{fig:supp_bpi}
\end{figure*}

\textbf{Where the Gains Come From.} \Cref{tab:supp_decomp} adds bootstrap intervals and the two BPI bounds to \cref{tab:decomp}, and \cref{fig:supp_bpi} plots BPI against the pairwise-residual control on every benchmark and the residual gain against $\Delta\BPI$. The intervals of the residual gain exclude zero on exactly the four datasets that pass Holm's correction, MUStARD, UR-FUNNY, SSW60, and AV-MNIST, which also have the largest $\Delta\BPI$.

\textbf{Sampling.} \Cref{tab:supp_sampling} compares independent draws with draws without replacement and with the positive removed, on the synthetic tasks together with the bank entries that share its discrete value. Accuracy barely changes, but the BPI readout needs independent draws. On the Gaussian control the bridge spreads over about 640 bank entries, far more than $K{=}64$, and both alternatives move BPI by less than 0.01 nats. On XOR-Mix it concentrates on about 9 entries, the bank copies of the two targets that pairs allow, so 64 draws without replacement must include targets that pairs reject, and BPI rises from 0.205 to 1.86 nats. Removing the positive as well identifies it by elimination, and BPI reaches 3.95 nats, close to $\log 65$.

\textbf{Bridge Capacity.} \Cref{tab:supp_bridgecap} extends \cref{tab:oracle} with $\BPI_{\text{pair}}$, the Gaussian control, and MUStARD. On the Gaussian control, whose continuous data also have $I^{(3)}=0$, a weaker bridge raises BPI to 0.21 nats while $\Delta\BPI$ stays below 0.01 nats. On MUStARD, where the oracle is unknown, BPI halves from the weakest to the strongest bridge and $\Delta\BPI$ changes by less than 0.01 nats.

\textbf{Rank and Cost.} With $R{=}128$ the retrieval embedding shrinks from 1024 to 640 dimensions (\cref{tab:supp_cost}) at a cost of 1.4\% on M3-Mix, the largest drop in \cref{fig:sensitivity}d.

\textbf{False Negatives.} \Cref{tab:supp_falseneg} measures how far valid alternatives are pushed below the positive. Top-$K$ mining suppresses them by 1.21 nats on M3-Mix, and decoys from the bridge by only 0.09, close to uniform negatives, which rarely sample them. Suppression is not zero, as expected from per-step gradients.

\begin{figure}[t]
  \centering
  \includegraphics[width=\linewidth]{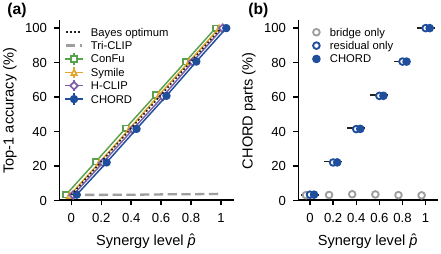}
  \caption{XOR task of Symile and ConFu. (a) Accuracy versus synergy level $\hat p$, mean $\pm$ std over three seeds, with the Bayes optimum dotted. (b) \method's bridge alone, residual alone, and full score, with the Bayes optimum as bars.}
  \label{fig:supp_xor}
\end{figure}

\textbf{Pure Synergy.} \Cref{fig:supp_xor} reports results on the XOR task over the synergy level $\hat p$, with three seeds per point (five at $\hat p=1$ in \cref{tab:controlled}). All methods use AdamW with learning rate $10^{-2}$, weight decay $10^{-2}$, batch 1000, and 100 epochs, and keep the checkpoint with the best validation accuracy. For \method, $K=32$, $R=16$, and the bank holds 1024 training targets. Since $I(a;b)=I(b;c)=0$, the bridge carries nothing about $b$ and its decoys are random bank targets, so \method reduces to a Hadamard critic trained against $K$ random negatives. It tracks the Bayes optimum within 1\% at every $\hat p$, as Symile, H-CLIP, and ConFu do, while Tri-CLIP stays at chance (a). The bridge alone stays at chance and the residual alone carries all of the accuracy (b). Baselines are our re-implementations under this protocol, and at $\hat p=1$ they match the published accuracy of Symile~\cite{saporta2024symile}.

\subsection{Training Together and Plugging In}
\label{supp:plugin}

\textbf{Training Variants.} The variants of \cref{fig:ablation} share the architecture, the bank, $K{=}32$, and the number of residual updates, and \cref{tab:supp_ablation} lists their values. The variant w/o Stop-grad uses the logits $b{+}r{-}\mathrm{sg}(b)$ in the decoy loss, where $\mathrm{sg}$ stops the gradient. Their value is $r$, so the loss and its optimum are those of \method, and only the gradient changes, which now also reaches the projection heads and $\beta$ through the bridge. Self-decoy draws the decoys from the bank softmax of $b{+}r$ and uses $b{+}r$ as logits, so the proposal contains the residual itself and no noise correction applies. Sequential trains $\Lpair+\Lcal$ alone for 5k steps, freezes the bridge, and then trains the residual for as many steps as \method.

\begin{figure}[t]
  \centering
  \includegraphics[width=\linewidth]{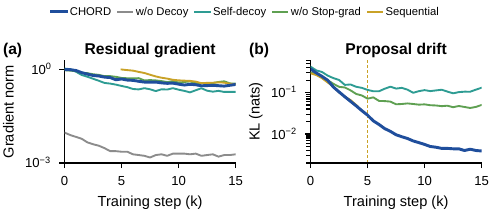}
  \caption{Training bridge and residual together on M3-Mix. (a) Normalized gradient norm reaching the residual. (b) Drift of the decoy proposal between steps 500 apart. Sequential follows \method until its bridge is frozen at 5k steps (dashed), after which the drift is zero.}
  \label{fig:supp_train}
\end{figure}

\Cref{fig:supp_train} shows the mechanism on M3-Mix. The gradient reaching the residual stays orders of magnitude lower without decoys. The drift of the proposal, $\KL(q^{t}\,\|\,q^{t+\Delta})$ between bank distributions 500 steps apart on fixed queries, vanishes for \method as the pairwise losses converge, but stays high when the decoy loss can move the proposal, through the bridge or by being part of it.

\begin{table}[t]
\centering
\caption{Values behind \cref{fig:ablation}, mean $\pm$ std over five seeds (\%, MUS: MUStARD).}
\label{tab:supp_ablation}
\footnotesize
\setlength{\tabcolsep}{2.4pt}
\begin{tabular}{@{}l cccc@{}}
\toprule
Variant & M3-Mix & MUS 2$\to$1 & MSR-VTT & MUS 1$\to$1 \\
\midrule
w/o Decoy & 66.9{\tiny$\pm$2.7} & 61.0{\tiny$\pm$3.6} & 47.4{\tiny$\pm$0.5} & 43.4{\tiny$\pm$1.5} \\
Full bank & 82.1{\tiny$\pm$1.2} & 62.3{\tiny$\pm$3.4} & 47.9{\tiny$\pm$0.5} & 43.3{\tiny$\pm$1.5} \\
Top-K & 79.6{\tiny$\pm$1.8} & 62.6{\tiny$\pm$3.3} & 46.9{\tiny$\pm$0.6} & 42.9{\tiny$\pm$1.6} \\
Reweighting & 71.3{\tiny$\pm$2.4} & 61.8{\tiny$\pm$3.5} & 47.3{\tiny$\pm$0.5} & 43.1{\tiny$\pm$1.5} \\
Self-decoy & 87.2{\tiny$\pm$1.5} & 62.7{\tiny$\pm$3.4} & 46.8{\tiny$\pm$0.6} & 41.4{\tiny$\pm$1.7} \\
w/o Stop-grad & 87.8{\tiny$\pm$1.1} & 63.3{\tiny$\pm$3.2} & 48.1{\tiny$\pm$0.4} & 41.9{\tiny$\pm$1.7} \\
Sequential & 89.3{\tiny$\pm$1.0} & 64.1{\tiny$\pm$3.0} & 48.3{\tiny$\pm$0.4} & 43.5{\tiny$\pm$1.4} \\
w/o Calib. & 84.0{\tiny$\pm$1.3} & 62.0{\tiny$\pm$3.3} & 47.2{\tiny$\pm$0.5} & 43.6{\tiny$\pm$1.4} \\
\midrule
\textbf{\method} & \textbf{89.6{\tiny$\pm$0.9}} & \textbf{64.4{\tiny$\pm$2.9}} & \textbf{48.4{\tiny$\pm$0.4}} & \textbf{43.6{\tiny$\pm$1.4}} \\
\bottomrule
\end{tabular}
\end{table}

\begin{table}[t]
\centering
\caption{The trained residual of \method added to other retrievers without retraining them ($\alpha{=}1$), with the gain over each retriever in small type. Bottom: mean gain over the eight retrievers for the residual and two controls. MUS: MUStARD, URF: UR-FUNNY (2$\to$1 R@10).}
\label{tab:supp_plugin_summary}
\footnotesize
\setlength{\tabcolsep}{2.4pt}
\begin{tabular}{@{}l ccccc@{}}
\toprule
Retriever & M3-Mix & MUS & URF & SSW60 & MSR-VTT \\
\midrule
Tri-CLIP & 72.9{\tiny\,+24.3} & 62.1{\tiny\,+3.1} & 13.9{\tiny\,+1.3} & 71.4{\tiny\,+1.1} & 47.2{\tiny\,+0.4} \\
GRAM & 71.0{\tiny\,+22.1} & 62.3{\tiny\,+2.1} & 4.3{\tiny\,+0.7} & 57.7{\tiny\,+0.6} & 47.6{\tiny\,+0.3} \\
TRIANGLE & 72.1{\tiny\,+22.9} & 52.4{\tiny\,+2.1} & 4.4{\tiny\,+0.8} & 65.0{\tiny\,+0.8} & 48.0{\tiny\,+0.2} \\
PMRL & 72.3{\tiny\,+23.5} & 59.8{\tiny\,+2.7} & 11.9{\tiny\,+1.1} & 67.1{\tiny\,+0.9} & 47.5{\tiny\,+0.4} \\
Symile & 80.4{\tiny\,+14.2} & 48.4{\tiny\,+1.9} & 12.8{\tiny\,+1.0} & 61.0{\tiny\,+0.6} & 45.2{\tiny\,+0.3} \\
Gated Symile & 81.6{\tiny\,+14.6} & 50.8{\tiny\,+1.5} & 13.1{\tiny\,+0.8} & 63.8{\tiny\,+0.7} & 45.8{\tiny\,+0.2} \\
H-CLIP & 82.1{\tiny\,+13.7} & 49.9{\tiny\,+1.8} & 13.0{\tiny\,+0.9} & 61.9{\tiny\,+0.9} & 45.5{\tiny\,+0.4} \\
ConFu & 75.8{\tiny\,+12.7} & 63.4{\tiny\,+0.7} & 14.1{\tiny\,+0.6} & 71.7{\tiny\,+0.5} & 47.4{\tiny\,+0.2} \\
\midrule
\multicolumn{6}{@{}l}{\emph{Mean gain over the eight retrievers}} \\[-1pt]
+ second seed & +0.7 & +0.6 & +0.3 & +0.3 & +0.3 \\
+ pairwise residual & +1.3 & +0.8 & +0.3 & +0.2 & +0.2 \\
\rowcolor{oursrow}+ residual & +18.5 & +2.0 & +0.9 & +0.8 & +0.3 \\
\bottomrule
\end{tabular}
\end{table}

\begin{table}[t]
\centering
\caption{Per-retriever results behind \cref{tab:supp_plugin_summary}, with the two controls and a weight $\alpha$ tuned on validation data. Small type: gain over the retriever alone.}
\label{tab:supp_plugin}
\scriptsize
\setlength{\tabcolsep}{1.6pt}
\begin{tabular}{@{}ll ccccc@{}}
\toprule
Retriever & Score & M3-Mix & MUS & URF & SSW60 & MSR-VTT \\
\midrule
\multirow{5}{*}{Tri-CLIP} & Retriever alone & 48.6 & 59.0 & 12.6 & 70.3 & 46.8 \\
 & + second seed & 49.0{\tiny\,+0.4} & 59.6{\tiny\,+0.6} & 12.8{\tiny\,+0.2} & 70.6{\tiny\,+0.3} & 47.1{\tiny\,+0.3} \\
 & + pairwise residual & 50.3{\tiny\,+1.7} & 60.1{\tiny\,+1.1} & 12.9{\tiny\,+0.3} & 70.6{\tiny\,+0.3} & 47.0{\tiny\,+0.2} \\
 & + residual & 72.9{\tiny\,+24.3} & 62.1{\tiny\,+3.1} & 13.9{\tiny\,+1.3} & 71.4{\tiny\,+1.1} & 47.2{\tiny\,+0.4} \\
 & + residual, tuned $\alpha$ & 73.8{\tiny\,+25.2} & 62.3{\tiny\,+3.3} & 14.0{\tiny\,+1.4} & 71.5{\tiny\,+1.2} & 47.3{\tiny\,+0.5} \\
\midrule
\multirow{5}{*}{GRAM} & Retriever alone & 48.9 & 60.2 & 3.6 & 57.1 & 47.3 \\
 & + second seed & 49.1{\tiny\,+0.2} & 60.5{\tiny\,+0.3} & 3.6{\tiny\,+0.0} & 57.3{\tiny\,+0.2} & 47.6{\tiny\,+0.3} \\
 & + pairwise residual & 50.6{\tiny\,+1.7} & 60.8{\tiny\,+0.6} & 3.8{\tiny\,+0.2} & 57.3{\tiny\,+0.2} & 47.5{\tiny\,+0.2} \\
 & + residual & 71.0{\tiny\,+22.1} & 62.3{\tiny\,+2.1} & 4.3{\tiny\,+0.7} & 57.7{\tiny\,+0.6} & 47.6{\tiny\,+0.3} \\
 & + residual, tuned $\alpha$ & 72.0{\tiny\,+23.1} & 62.5{\tiny\,+2.3} & 4.4{\tiny\,+0.8} & 57.9{\tiny\,+0.8} & 47.6{\tiny\,+0.3} \\
\midrule
\multirow{5}{*}{TRIANGLE} & Retriever alone & 49.2 & 50.3 & 3.6 & 64.2 & 47.8 \\
 & + second seed & 49.5{\tiny\,+0.3} & 50.9{\tiny\,+0.6} & 3.9{\tiny\,+0.3} & 64.5{\tiny\,+0.3} & 48.1{\tiny\,+0.3} \\
 & + pairwise residual & 50.7{\tiny\,+1.5} & 51.0{\tiny\,+0.7} & 4.0{\tiny\,+0.4} & 64.4{\tiny\,+0.2} & 47.9{\tiny\,+0.1} \\
 & + residual & 72.1{\tiny\,+22.9} & 52.4{\tiny\,+2.1} & 4.4{\tiny\,+0.8} & 65.0{\tiny\,+0.8} & 48.0{\tiny\,+0.2} \\
 & + residual, tuned $\alpha$ & 73.0{\tiny\,+23.8} & 52.6{\tiny\,+2.3} & 4.6{\tiny\,+1.0} & 65.1{\tiny\,+0.9} & 48.1{\tiny\,+0.3} \\
\midrule
\multirow{5}{*}{PMRL} & Retriever alone & 48.8 & 57.1 & 10.8 & 66.2 & 47.1 \\
 & + second seed & 49.1{\tiny\,+0.3} & 57.8{\tiny\,+0.7} & 11.1{\tiny\,+0.3} & 66.6{\tiny\,+0.4} & 47.4{\tiny\,+0.3} \\
 & + pairwise residual & 50.1{\tiny\,+1.3} & 58.0{\tiny\,+0.9} & 11.1{\tiny\,+0.3} & 66.5{\tiny\,+0.3} & 47.3{\tiny\,+0.2} \\
 & + residual & 72.3{\tiny\,+23.5} & 59.8{\tiny\,+2.7} & 11.9{\tiny\,+1.1} & 67.1{\tiny\,+0.9} & 47.5{\tiny\,+0.4} \\
 & + residual, tuned $\alpha$ & 73.2{\tiny\,+24.4} & 60.0{\tiny\,+2.9} & 12.0{\tiny\,+1.2} & 67.2{\tiny\,+1.0} & 47.5{\tiny\,+0.4} \\
\midrule
\multirow{5}{*}{Symile} & Retriever alone & 66.2 & 46.5 & 11.8 & 60.4 & 44.9 \\
 & + second seed & 67.4{\tiny\,+1.2} & 47.3{\tiny\,+0.8} & 12.2{\tiny\,+0.4} & 60.7{\tiny\,+0.3} & 45.2{\tiny\,+0.3} \\
 & + pairwise residual & 67.5{\tiny\,+1.3} & 47.4{\tiny\,+0.9} & 12.2{\tiny\,+0.4} & 60.5{\tiny\,+0.1} & 45.1{\tiny\,+0.2} \\
 & + residual & 80.4{\tiny\,+14.2} & 48.4{\tiny\,+1.9} & 12.8{\tiny\,+1.0} & 61.0{\tiny\,+0.6} & 45.2{\tiny\,+0.3} \\
 & + residual, tuned $\alpha$ & 81.9{\tiny\,+15.7} & 48.9{\tiny\,+2.4} & 13.0{\tiny\,+1.2} & 61.3{\tiny\,+0.9} & 45.3{\tiny\,+0.4} \\
\midrule
\multirow{5}{*}{Gated Symile} & Retriever alone & 67.0 & 49.3 & 12.3 & 63.1 & 45.6 \\
 & + second seed & 68.1{\tiny\,+1.1} & 49.9{\tiny\,+0.6} & 12.6{\tiny\,+0.3} & 63.5{\tiny\,+0.4} & 45.9{\tiny\,+0.3} \\
 & + pairwise residual & 68.2{\tiny\,+1.2} & 50.0{\tiny\,+0.7} & 12.5{\tiny\,+0.2} & 63.4{\tiny\,+0.3} & 45.8{\tiny\,+0.2} \\
 & + residual & 81.6{\tiny\,+14.6} & 50.8{\tiny\,+1.5} & 13.1{\tiny\,+0.8} & 63.8{\tiny\,+0.7} & 45.8{\tiny\,+0.2} \\
 & + residual, tuned $\alpha$ & 82.8{\tiny\,+15.8} & 51.2{\tiny\,+1.9} & 13.3{\tiny\,+1.0} & 64.0{\tiny\,+0.9} & 45.9{\tiny\,+0.3} \\
\midrule
\multirow{5}{*}{H-CLIP} & Retriever alone & 68.4 & 48.1 & 12.1 & 61.0 & 45.1 \\
 & + second seed & 69.6{\tiny\,+1.2} & 48.8{\tiny\,+0.7} & 12.4{\tiny\,+0.3} & 61.5{\tiny\,+0.5} & 45.4{\tiny\,+0.3} \\
 & + pairwise residual & 69.5{\tiny\,+1.1} & 48.9{\tiny\,+0.8} & 12.3{\tiny\,+0.2} & 61.2{\tiny\,+0.2} & 45.3{\tiny\,+0.2} \\
 & + residual & 82.1{\tiny\,+13.7} & 49.9{\tiny\,+1.8} & 13.0{\tiny\,+0.9} & 61.9{\tiny\,+0.9} & 45.5{\tiny\,+0.4} \\
 & + residual, tuned $\alpha$ & 83.4{\tiny\,+15.0} & 50.3{\tiny\,+2.2} & 13.2{\tiny\,+1.1} & 62.1{\tiny\,+1.1} & 45.5{\tiny\,+0.4} \\
\midrule
\multirow{5}{*}{ConFu} & Retriever alone & 63.1 & 62.7 & 13.5 & 71.2 & 47.2 \\
 & + second seed & 64.3{\tiny\,+1.2} & 63.1{\tiny\,+0.4} & 13.9{\tiny\,+0.4} & 71.5{\tiny\,+0.3} & 47.5{\tiny\,+0.3} \\
 & + pairwise residual & 64.1{\tiny\,+1.0} & 63.2{\tiny\,+0.5} & 13.8{\tiny\,+0.3} & 71.4{\tiny\,+0.2} & 47.4{\tiny\,+0.2} \\
 & + residual & 75.8{\tiny\,+12.7} & 63.4{\tiny\,+0.7} & 14.1{\tiny\,+0.6} & 71.7{\tiny\,+0.5} & 47.4{\tiny\,+0.2} \\
 & + residual, tuned $\alpha$ & 77.6{\tiny\,+14.5} & 63.8{\tiny\,+1.1} & 14.2{\tiny\,+0.7} & 71.8{\tiny\,+0.6} & 47.5{\tiny\,+0.3} \\
\bottomrule
\end{tabular}
\end{table}

\begin{table}[t]
\centering
\caption{The residual added to other retrievers ($\alpha{=}1$) on the remaining two-to-one benchmarks. YC2 uses the residual trained on MSR-VTT. Small type: gain over the retriever alone.}
\label{tab:supp_plugin_other}
\scriptsize
\setlength{\tabcolsep}{1.8pt}
\begin{tabular}{@{}l cccccc@{}}
\toprule
Retriever & MOSI & MOSEI & AV-MNIST & VB100 & VATEX & YC2 \\
\midrule
Tri-CLIP & 23.4{\tiny\,+0.3} & 9.1{\tiny\,+0.2} & 70.6{\tiny\,+1.2} & 19.5{\tiny\,+0.3} & 70.8{\tiny\,+0.4} & 12.6{\tiny\,+0.0} \\
GRAM & 12.4{\tiny\,+0.7} & 5.3{\tiny\,+0.1} & 65.4{\tiny\,+1.3} & 8.3{\tiny\,+0.4} & 71.6{\tiny\,+0.4} & 11.5{\tiny\,+0.1} \\
TRIANGLE & 7.6{\tiny\,+0.1} & 6.0{\tiny\,+0.2} & 66.0{\tiny\,+0.9} & 12.3{\tiny\,+0.5} & 72.1{\tiny\,+0.3} & 10.7{\tiny\,+0.0} \\
PMRL & 21.9{\tiny\,+0.3} & 8.0{\tiny\,+0.2} & 67.1{\tiny\,+1.1} & 15.4{\tiny\,+0.3} & 71.4{\tiny\,+0.4} & 11.8{\tiny\,+0.0} \\
Symile & 17.4{\tiny\,+0.4} & 7.2{\tiny\,+0.1} & 71.4{\tiny\,+0.4} & 13.6{\tiny\,+0.5} & 68.2{\tiny\,+0.3} & 10.8{\tiny\,$-$0.1} \\
Gated Symile & 18.5{\tiny\,+0.1} & 7.7{\tiny\,+0.1} & 71.8{\tiny\,+0.3} & 17.3{\tiny\,+0.1} & 69.1{\tiny\,+0.3} & 11.3{\tiny\,+0.1} \\
H-CLIP & 18.3{\tiny\,+0.2} & 7.6{\tiny\,+0.2} & 71.7{\tiny\,+0.4} & 14.2{\tiny\,+0.3} & 68.3{\tiny\,+0.2} & 10.7{\tiny\,$-$0.1} \\
ConFu & 19.9{\tiny\,+0.1} & 8.5{\tiny\,+0.1} & 71.6{\tiny\,+0.7} & 18.3{\tiny\,+0.0} & 71.5{\tiny\,+0.2} & 11.6{\tiny\,+0.0} \\
\bottomrule
\end{tabular}
\end{table}

\begin{table}[t]
\centering
\caption{Bridge as a first stage. Only the top-$k$ candidates of the bridge receive the residual (T$\to$VA R@1, \%). The last column scores every candidate.}
\label{tab:supp_cascade}
\footnotesize
\setlength{\tabcolsep}{5pt}
\begin{tabular}{@{}l cccc@{}}
\toprule
 & $k{=}10$ & $k{=}50$ & $k{=}100$ & All \\
\midrule
MSR-VTT & 47.9 & 48.3 & 48.4 & 48.4 \\
VATEX & 71.8 & 72.3 & 72.4 & 72.4 \\
\bottomrule
\end{tabular}
\end{table}

\textbf{The Residual as a Plug-In.} The residual reads only the inputs, so it can be added to the score of another retriever without retraining that retriever. Each retriever is trained as in \cref{tab:controlled,tab:multibench,tab:audiovisual} and scores a candidate by its logit divided by its own learned temperature, which, like the residual, estimates a log-ratio in nats. The residual comes from the trained \method model of the same dataset and seed and is computed from its own features, which on the LanguageBind benchmarks are the frozen encoders that all retrievers share. The two scores are added with $\alpha{=}1$, and a tuned variant picks $\alpha\in\{0.25,0.5,1,2\}$ on validation data. Two controls separate the gain from ensembling. One adds the score of the same retriever trained with another seed, and the other adds a pairwise residual $\sum_{i<j}\vw_{ij}^\top(\va_i\odot\va_j)$ of rank $R$ trained on the same decoys.

As shown in \cref{tab:supp_plugin_summary}, the residual lifts all eight retrievers on M3-Mix by a wide margin, and the product critics, whose own scores already carry part of the interaction, reach the highest accuracy. On real data the mean gains follow $\Delta\BPI$, as the Plug-in column of \cref{tab:decomp} shows, and both controls add far less where $\Delta\BPI$ is large. On MSR-VTT, where $\Delta\BPI$ is small, the residual adds no more than a second seed. ConFu gains least, since a residual fitted to the misfit of our bridge transfers best to retrievers close to that bridge. \Cref{tab:supp_plugin} gives every retriever and control, and the means of \cref{tab:supp_plugin_summary} average its rows. The Plug-in column of \cref{tab:decomp} averages \cref{tab:supp_plugin,tab:supp_plugin_other} over the eight retrievers. A tuned $\alpha$ adds little, most for the product critics, whose own scores already carry part of the interaction. On the remaining datasets in \cref{tab:supp_plugin_other}, gains are small where $\Delta\BPI$ is small and vanish on YouCook2, as for \method itself.

\textbf{Bridge as a First Stage.} The bridge can also preselect candidates, with the residual scoring only its top $k$. \Cref{tab:supp_cascade} shows that $k{=}100$ already matches full scoring on MSR-VTT and VATEX. Full scoring costs one inner product over $d{+}R$ dimensions per candidate, so the cascade is an option for very large galleries rather than part of the method.

\subsection{More than Three Modalities}
\label{supp:mmod}

With $M$ modalities, the bridge for target $k$ sums $\beta_{ik}\vu_i$ over all other modalities, and one residual of the form \cref{eq:residual} is shared across all triples that contain $k$, so the residual score sums over these triples. Decoys are drawn from the bridge conditioned on all $M{-}1$ observed modalities. \Cref{tab:scaling} (right) reports the results on Vision\&Touch. Symile and H-CLIP degrade as $M$ grows because a single product over all modalities becomes harder to train from the same negatives.

A sum of triple products covers interactions up to order three only. \Cref{tab:supp_parity} makes this boundary explicit with four modalities. When $x_4=x_1\oplus x_2$, the triple residual solves the task. When $x_4=x_1\oplus x_2\oplus x_3$, every pair and every triple is independent, and the triple residual stays at chance, like Tri-CLIP. Adding a four-way product $\vw_4^\top(\va_1\odot\va_2\odot\va_3\odot\va_4)$ recovers the task, as does the four-way product of Symile. Parity needs no new decoys because its lower-order terms vanish. In general, a $k$-way term would need decoys that fit all terms of lower order, drawn from the bridge together with the lower-order residuals.

\begin{table}[t]
\centering
\caption{Parity with four modalities (top-1 \%, 32 candidates, chance 3.1). Third-order: $x_4=x_1\oplus x_2$ with $x_3$ independent. Four-way: $x_4=x_1\oplus x_2\oplus x_3$, where every pair and every triple is independent.}
\label{tab:supp_parity}
\footnotesize
\setlength{\tabcolsep}{4pt}
\begin{tabular}{@{}l cc@{}}
\toprule
Method & Third-order & Four-way \\
\midrule
Tri-CLIP & 3.2 & 3.1 \\
Symile (four-way product) & 99.1 & 99.4 \\
\rowcolor{oursrow}\method (triple residuals) & 100.0 & 3.3 \\
\method + four-way term & 99.8 & 99.6 \\
\bottomrule
\end{tabular}
\end{table}

\begin{figure}[t]
  \centering
  \includegraphics[width=\linewidth]{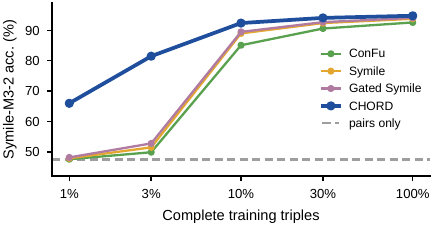}
  \caption{Symile-M3-2 accuracy when only a share of the training tuples is complete (full curve of \cref{tab:scaling}, left).}
  \label{fig:supp_eff}
\end{figure}

\begin{table}[t]
\centering
\caption{Two-to-one top-1 accuracy (\%) with 15 uniformly drawn candidates (Rand@16) and with 15 decoys
from the held-out judge bridge (Decoy@16).}
\label{tab:supp_probe_sanity}
\footnotesize
\setlength{\tabcolsep}{2.6pt}
\begin{tabular}{@{}l cc cc cc@{}}
\toprule
 & \multicolumn{2}{c}{MOSI} & \multicolumn{2}{c}{UR-FUNNY} & \multicolumn{2}{c}{MUStARD} \\
\cmidrule(lr){2-3}\cmidrule(lr){4-5}\cmidrule(l){6-7}
Method & Rand & Decoy & Rand & Decoy & Rand & Decoy \\
\midrule
Tri-CLIP & \underline{61.8} & 8.4 & 43.5 & 7.7 & 78.6 & 9.9 \\
Symile & 55.3 & \underline{9.6} & 41.2 & \underline{9.3} & 70.1 & \underline{13.1} \\
ConFu & 59.4 & 9.3 & \underline{44.0} & 9.0 & \underline{80.2} & 12.0 \\
\rowcolor{oursrow}\textbf{CHORD} & \textbf{63.0} & \textbf{11.9} & \textbf{45.1} & \textbf{14.4} & \textbf{81.9} & \textbf{18.3} \\
\bottomrule
\end{tabular}
\end{table}

\begin{figure}[t]
  \centering
  \includegraphics[width=\linewidth]{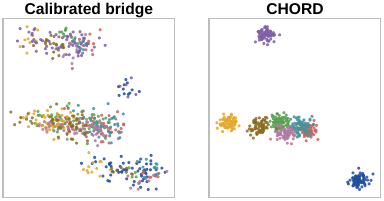}
  \caption{t-SNE of two-to-one query embeddings on M3-Mix test samples that need the speech, colored by the target class. The bridge places a query between the two classes named in its caption, whereas \method separates queries by their target.}
  \label{fig:supp_tsne}
\end{figure}

\textbf{Qualitative View.} \Cref{fig:supp_tsne} embeds the two-to-one queries of \method, the concatenation used in \cref{eq:score}, and those of its bridge alone. \Cref{fig:supp_eff} gives the full data-efficiency curve. Without the residual, queries cluster by the pair of classes in the caption, and with it they cluster by the class the speech selects.

\end{document}